\pdfoutput=1
\documentclass[11pt]{article}

\usepackage[]{acl}

\usepackage{times}
\usepackage{latexsym}
\usepackage{url}

\usepackage{graphicx}
\usepackage{adjustbox}
\usepackage{booktabs}
\usepackage{colortbl}
\usepackage{subfigure}
\usepackage{amsmath}
\usepackage{amssymb}
\usepackage{pifont}

\newtheorem{definition}{Definition}
\newtheorem{property}{Property}
\newtheorem{remark}{Remark}

\usepackage[T1]{fontenc}
\usepackage[utf8]{inputenc}

\usepackage{microtype}

\usepackage{inconsolata}

\title{Minimal Distillation Schedule for Extreme\\ Language Model Compression}

\author{Chen Zhang\textsuperscript{\ding{168}}, Yang Yang\textsuperscript{\ding{169}}, Qifan Wang\textsuperscript{\ding{170}}, Jiahao Liu\textsuperscript{\ding{169}}, Jingang Wang\textsuperscript{\ding{169}}, \\
  \textbf{Wei Wu\textsuperscript{\ding{169}}, Dawei Song\textsuperscript{\ding{168}}\Thanks{ Dawei Song is the corresponding author.}} \\
  \textsuperscript{\ding{168}}Beijing Institute of Technology \quad
  \textsuperscript{\ding{169}}Meituan NLP \quad
  \textsuperscript{\ding{170}}Meta AI \\
  \texttt{chenzhang9702@outlook.com}}
  
\begin{document}

\maketitle

\begin{abstract}
%Recent studies have uncovered that language model distillation is less effective when facing a large capacity gap between the teacher and the student, and introduced teacher assistant-based distillation to bridge the gap. 
%As a connection, the scale and the performance of the teacher assistant is of vital importance to bring the knowledge from the teacher to the student. 
%However, existing teacher assistant-based methods require maximally many trials before scheduling an optimal teacher assistant.
% To this end, we propose a \underline{mini}mal \underline{di}stillation \underline{sc}hedule (\textsc{MiniDisc}) for scheduling the optimal teacher assistant in minimally one trial.
% In particular, motivated by the finding that the performance of the student is positively correlated to the scale-performance tradeoff of the teacher assistant, \textsc{MiniDisc} is designed with a \textit{$\lambda$-tradeoff} to measure the optimality of the teacher assistant without trial distillation to the student. \textsc{MiniDisc} then can schedule the optimal teacher assistant with the best \textit{$\lambda$-tradeoff} in a \textit{sandwich framework}. \textsc{MiniDisc} is evaluated with an extensive set of experiments on GLUE. Experimental results demonstrate the improved efficiency our \textsc{MiniDisc} compared to several state-of-the-art baselines. We further apply \textsc{MiniDisc} to a language model with billions of parameters and show its scalability.%\footnote{Code is available at \url{https://github.com/GeneZC/MiniDisc}}. %We will release our code and scripts for exact reproducibility.
Recent studies have revealed that language model distillation is less effective when there is a significant capacity gap between the teacher and the student models. In order to address this issue, teacher assistant-based distillation has been introduced to bridge the gap. The selection of the teacher assistant plays a crucial role in transferring knowledge from the teacher to the student. However, existing approaches for teacher assistant-based distillation require numerous trials to find the optimal teacher assistant.
In this paper, we propose a novel approach called \underline{Mini}mal \underline{Di}stillation \underline{Sc}hedule (\textsc{MiniDisc}), which enables the scheduling of an optimal teacher assistant in just one trial for extreme model compression (e.g, to 5\% scale). In particular, we empirically show that the performance of the student is positively correlated with the scale-performance tradeoff of the teacher assistant. We then introduce a new $\lambda$-tradeoff metric that quantifies the optimality of the teacher assistant without the need for trial distillation to the student. By employing a sandwich framework, \textsc{MiniDisc} can select the optimal teacher assistant with the best $\lambda$-tradeoff.
We extensively evaluate \textsc{MiniDisc} through a series of experiments on the GLUE benchmark. The experimental results demonstrate the improved efficiency of our approach compared to several state-of-the-art baselines. Furthermore, we showcase the scalability of \textsc{MiniDisc} by applying it to a language model with billions of parameters.
\end{abstract}

\section{Introduction}

\begin{figure}[ht]
    \centering
    \includegraphics[width=0.49\textwidth]{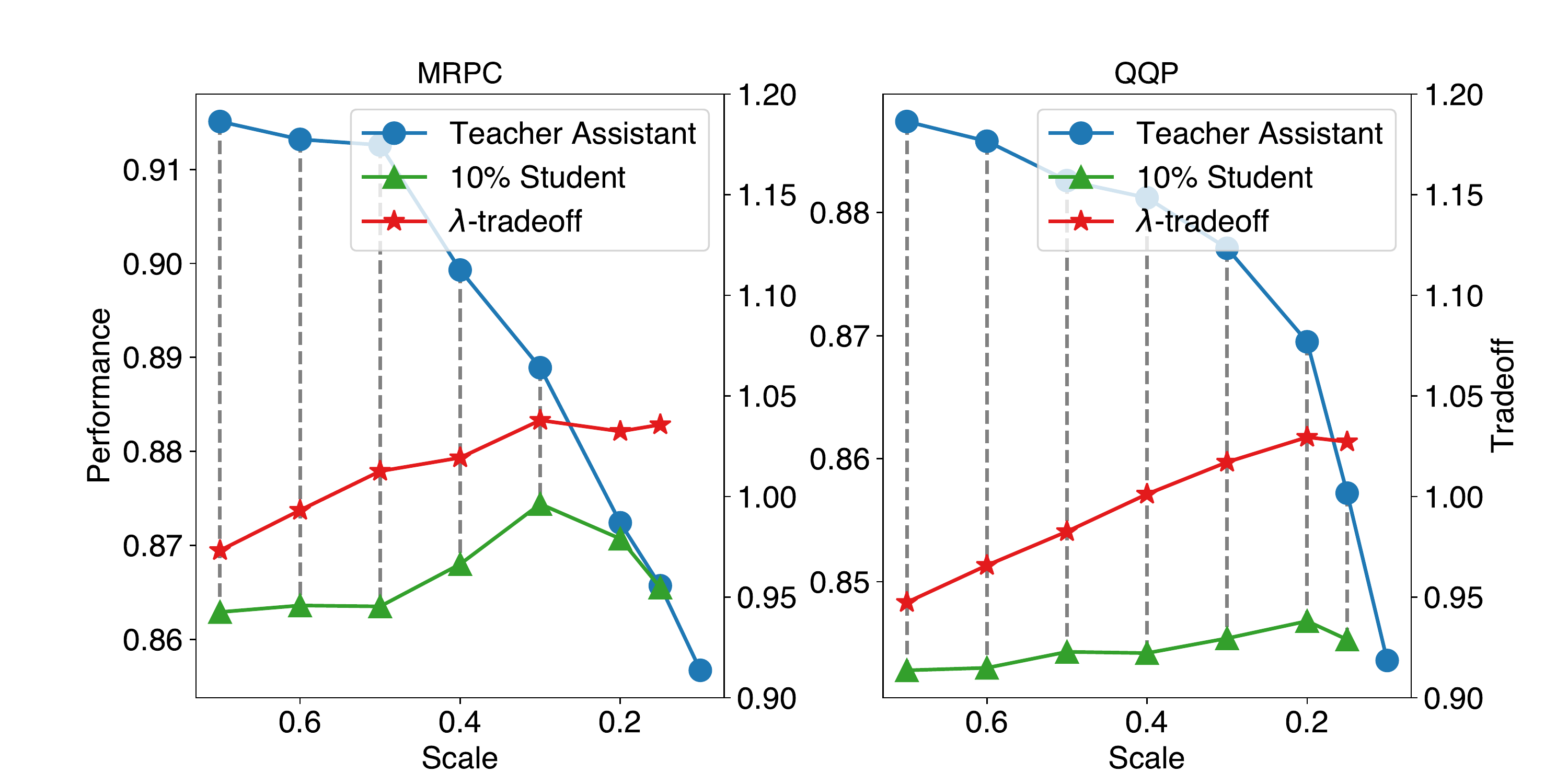}
    \caption{The impact of teacher assistants of different scales and performance on the performance of students. In the study, a BERT\textsubscript{\sf base} model is used as the teacher and distilled to a pruned student (10\% parameters of the teacher) via different teacher assistants~\citep{MirzadehFLLMG20} on MRPC and QQP. There are several observations: (1) The \textcolor[rgb]{0.12,0.47,0.71}{blue} curve shows that the performance of the teacher assistant degrades with the decreasing of its scale, which is obvious. (2) The \textcolor[rgb]{0.2,0.63,0.17}{green} curve validates that the performance of the student varies with different teacher assistants. (3) The \textcolor[rgb]{0.89,0.10,0.11}{red} curve represents $\lambda$-tradeoff of the teacher assistant, which is positively correlated with the performance of the student.
    }
    \label{fig:1}
    \vspace{-5mm}
\end{figure}

Pretrained language models (LMs)~\citep{DevlinCLT19,LiuOG19,Radford19,BrownMRSKDNSSAA20,RaffelSRLNMZLL20} have achieved promising results in various downstream tasks~\citep{WangSMHLB19,RajpurkarJL18}, but are inapplicable to those requiring limited computational resources~\citep{LiuSH21}. To address this issue, LMs can be compressed using a range of strategies such as model quantization~\citep{ZafrirBIW19,BaiZHSJJLLK20}, pruning~\citep{MichelLN19,HouHSJCL20}, etc., among which knowledge distillation~\citep{SunCGL19,WangW0B0020} has gained significant attention. It operates within the teacher-student framework, where a large model acts as the teacher, transferring its knowledge to a smaller student model.

Recent advances~\citep{MirzadehFLLMG20} have shown a significant performance decline in conventional distillation methods when dealing with a substantial capacity gap between the teacher and the student models. 
To alleviate this, teacher assistant-based distillation~\citep{SonNCH21} has been proposed. This approach involves distilling the teacher model into an intermediate-scale teacher assistant, which then serves as an intermediary to transfer knowledge to the student model.
While teacher assistant-based distillation generally lifts the performance of the student~\citep{WangW0B0020,WuRGH021}, the performance of the student is largely impacted by the choice of the teacher assistant as illustrated in Figure~\ref{fig:1}. In fact, we observe there is potentially a turning point of the student performance, indicating a scale-performance (i.e., x- v.s. y-axis) tradeoff in scheduling the teacher assistant.
However, existing studies schedule the teacher assistant in an enumeration manner, resulting in an inferior solution that requires maximally many trials to meet the optimal teacher assistant (maximal distillation schedule, in short \textsc{MaxiDisc}).

To this demand, we propose a minimal distillation schedule (\textsc{MiniDisc}) that enables the identification of the optimal teacher assistant in just a single trial. We define a \textit{$\lambda$-tradeoff} metric to empirically measure the tradeoff between scale and performance for a given teacher assistant, as depicted in Figure~\ref{fig:1}. This allows us to determine the optimality of the teacher assistant without requiring multiple trial distillations to the student model.
To efficiently obtain the optimal teacher assistant based on the \textit{$\lambda$-tradeoff} metric, we introduce \textsc{MiniDisc} within a sandwich framework, consisting of three stages. In the \textit{specification} stage, we utilize gridding and pruning techniques to generate a series of teacher assistant candidates with varying scales.
In the \textit{optimization} stage, we demonstrate that the generated candidates adhere to the incremental property and the sandwich rule. Furthermore, we present two approximations that enable the computation of the \textit{$\lambda$-tradeoff} for each teacher assistant candidate at a lower computational cost.
In the \textit{selection} stage, we choose the optimal teacher assistant by selecting the candidate with the highest $\lambda$-tradeoff value.
It is worth noting that \textsc{MiniDisc} can be directly extended to scenarios involving multiple sequential teacher assistants by recursively applying the \textsc{MiniDisc} procedure. However, this work focuses on a single teacher assistant as it is sufficiently effective.
% that schedules the optimal teacher assistant in only one trial.
% Specifically, \textit{$\lambda$-tradeoff} is defined to empirically measure the scale-performance tradeoff for any given teacher assistant as in Figure~\ref{fig:1} so that the optimality of the teacher assistant can be determined without trial distillation to the student.
% In order to efficiently obtain the optimal teacher assistant according to \textit{$\lambda$-tradeoff}, \textsc{MiniDisc} is implemented in a \textit{sandwich framework}, which includes three stages.
% In \textit{specification}, we leverage gridding and pruning techniques~\citep{LiKDSG17,FrankleC19} to generate a series of teacher assistant candidates of different scales. 
% In \textit{optimization}, we demonstrate that the generated candidates satisfy the \textit{incremental property} and the \textit{sandwich rule}, and present two approximations that yields the \textit{$\lambda$-tradeoff} of the teacher assistant candidate of each scale in low compute.
% In \textit{selection}, we select the optimal teacher assistant with the largest \textit{$\lambda$-tradeoff} value.
% It is noteworthy that \textsc{MiniDisc} can be flexibly extended to the case with multiple sequential teacher assistants by recursively applying \textsc{MiniDisc}. However, this work only focuses on one teacher assistant between the teacher and student due to its sufficiency. 

To verify the effectiveness of \textsc{MiniDisc}, we conduct experiments on GLUE~\citep{WangSMHLB19}.
Experimental results exhibit the competitive performance of \textsc{MiniDisc} compared to several state-of-the-art baselines, with improved efficiency (10$\times$) of \textsc{MiniDisc} compared to \textsc{MaxiDisc}. 
Further, \textsc{MiniDisc} is applied to large LMs EncT5\textsubscript{\sf xl}~\citep{LiuSY21} and LLaMA2\textsubscript{\sf 7B}~\citep{Touvron23} to show its scalability.

% Our main contributions are summarized as follows:
% \begin{itemize}
% \item  We investigate the impact of teacher assistants with different scales on the performance of the student, and introduce a quantitative scale-performance tradeoff measure, \textit{$\lambda$-tradeoff}, on the teacher assistant that is positively correlated with the student performance.

% \item We show two properties of the specified candidates. These properties lead to a novel optimization framework, \textit{sandwich framework}, that efficiently achieves yielding \textit{$\lambda$-tradeoff}s of teacher assistants of all scales. \textit{sandwich framework}, together with the \textit{$\lambda$-tradeoff}, enables a minimal distillation schedule. 

% \item We validate the effectiveness and efficiency of the \textsc{MiniDisc}. Our results of a LM with over one billion parameters show the scalability of our approach. To our best knowledge, our work is the first one exploring the distillation of large LMs.
% \end{itemize}
\section{Related Work}

% \paragraph{Language Model}

% Language models (LMs)~\citep{DevlinCLT19,RaffelSRLNMZLL20} are widely adopted in various natural language tasks~\citep{BaoHWWW20,ZhangZSL20}. Typical LMs consist of a stack of transformer~\citep{VaswaniSPUJGKP17} encoder/decoder layers. Each encoder layer has two modules. The first is a self-attention module, and the second is a feed-forward module. A residual connection is employed around each of these modules, with a layer normalization placed either in (pre-norm) or out of (post-norm) the connection~\citep{XiongYHZZXZLWL20}. Each decoder layer additionally has a cross-attention module between the self-attention and feed-forward modules. While LMs exhibit excellent performance in various downstream tasks, their scales impede the deployment in real-world applications. Therefore, it is an important research problem of learning compact language models from the large ones.
% In our work, we aim to make LMs deployable via model compression. 

\paragraph{Model Pruning}

%Inspired by the idea that not all parameters contribute equally to the overall performance of a model, model pruning is widely adopted to waive the parameters with little impact. 
Model pruning~\citep{HanPTD15} spans from unstructured pruning~\citep{FrankleC19,LouizosWK18,Sanh0R20,ChenFC0ZWC20} to structured pruning~\citep{MichelLN19,HouHSJCL20,LiKDSG17,XiaZC22,LagunasCSR21}. Unstructured pruning prunes parameters at neuron level referring to parameter magnitude~\citep{HanPTD15,LouizosWK18} or learning dynamics~\citep{Sanh0R20}, while structured pruning~\citep{MichelLN19,XiaZC22} prunes parameters at module level relying on parameter sensitivity. 
Although unstructured pruning enjoys a finer-grained pruning, it can only fit specialized devices. In contrast, structured pruning generally fits modern acceleration devices. In our work, we adopt structured pruning for deriving the structures of candidates for its benefits for distillation. Pruning also offers an opportunity to optimize the efficiency and effectiveness of our method due to its merits~\citep{LiKDSG17,FrankleC19,YuH19,CaiGWZH20,LiangZCJLHZC20,Ma22,YangZWS22,Yang22}.

\paragraph{Knowledge Distillation}

%Knowledge distillation is employed as a promising choice for model compression. 
Knowledge distillation~\citep{HintonVD15} can be divided into two categories: task-specific~\citep{SunCGL19,HintonVD15,LiLZXYJ20,ParkKY21} and task-agnostic~\citep{WangW0B0020,TurcCLT19,SanhDCW19,SunYSLYZ20,JiaoYSJCL0L20,WangBHDW21} distillation. Task-specific methods distill finetuned models with task-specific data, while task-agnostic methods distill pretrained models directly with task-agnostic data. 
Learning bjective is central to distillation, and distilling logits~\citep{HintonVD15} is the most common way. 
Recently, hidden states~\citep{SanhDCW19,SunYSLYZ20}, attention distributions~\citep{JiaoYSJCL0L20,WangW0B0020,LiLZXYJ20,WangBHDW21}, and high-order relations~\citep{ParkKY21} are taken into consideration for better abstraction. Teacher assistant-based distillation~\citep{WangW0B0020,MirzadehFLLMG20,WuRGH021} is showcased to trade in teacher scale for student performance by inserting an intermediate teacher assistant. 
%This phenomenon is also supported in other work that better student performance should be attained with slightly worse teacher learning capacity~\citep{ZhouXM21}. 
However, setting an optimal teacher assistant for the student is nontrivial. In this work, we aim to achieve this goal.

\section{Methodology}

\begin{figure}
    \centering
    \includegraphics[width=0.45\textwidth]{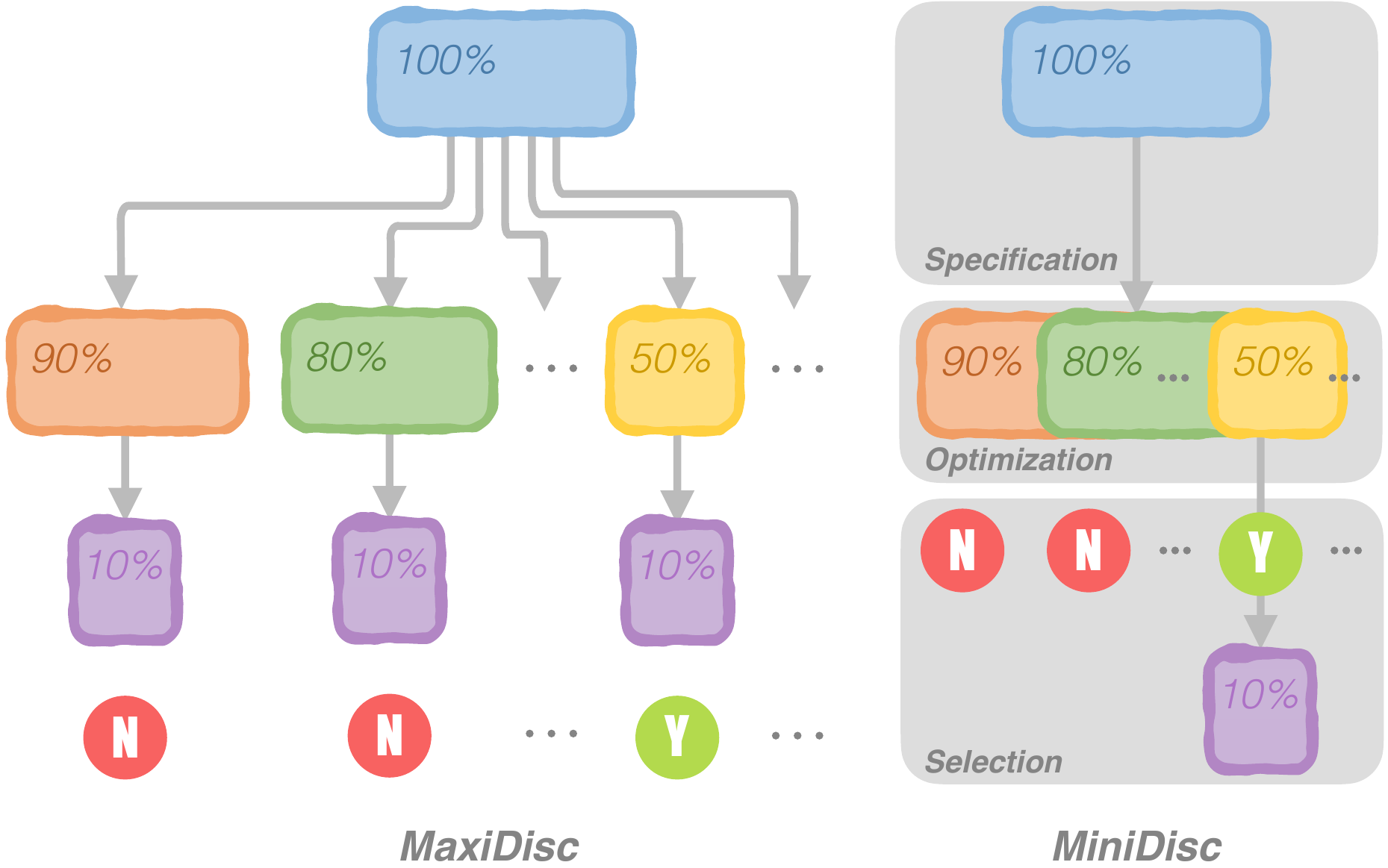}
    \caption{An overview of \textsc{MiniDisc} by contrasting it to \textsc{MaxiDisc}, where one arrow denotes a distillation step. \textsc{MiniDisc} uses only one trial while \textsc{MaxiDisc} uses many trials to schedule the optimal teacher assistant. %\textit{Specification}: the scales and structures of candidates are specified by gridding the scale and pruning the structure of the teacher. \textit{Optimization}: candidates are sub-sampled and assembled into a sandwich-like model, thus optimized in the \textit{sandwich framework}. \textit{Selection}: the candidate with the best \textit{$\lambda$-tradeoff} is selected, thus the student is distilled.
    }
    \label{fig:2}
    \vspace{-5mm}
\end{figure}

\subsection{Problem Definition}

Given a teacher model $\mathcal{T}$, our goal is to identify an optimal teacher assistant $\mathcal{A}$, such that the performance of the student $\mathcal{S}$ can be maximized when distilling the teacher to the student via the teacher assistant (i.e., $\mathcal{T}\rightarrow\mathcal{A}\rightarrow\mathcal{S}$). Formally, the teacher model is denoted as $(\mathcal{T}, s_t, m_t)$, where $s_t$ and $m_t$ are the \textbf{s}acle and perfor\textbf{m}ance of the teacher respectively. Similarly, the teacher assistant and the student are denoted as $(\mathcal{A}, s_a, m_a)$ and $(\mathcal{S}, s_s, m_s)$
% that should be distilled to a student $(\mathcal{S}, s_s, m_s$), the goal is to find a teacher assistant $(\mathcal{A}, s_a, m_a)$ such that the student performance can be maximized when distilling the teacher to the student via the teacher assistant (i.e., $\mathcal{T}\rightarrow\mathcal{A}\rightarrow\mathcal{S}$). Here \textbf{the first, second and third elements} in a tuple denote \textbf{the structure, the scale, and the performance} respectively. 
It is straightforward that the scale and the performance of the teacher assistant are bounded by the teacher and the student.
%, i.e., $s_s\le s_a\le s_t$ and $m_s\le m_a\le m_t$.

The overview of \textsc{MiniDisc} is presented in Figure~\ref{fig:2}. Our \textsc{MiniDisc} uses only one trial while \textsc{MaxiDisc} uses many trials to schedule the optimal teacher assistant. There are three key components in \textsc{MiniDisc}. \textit{Specification}: the scales and structures of candidates are specified by gridding the scale and pruning the structure of the teacher. \textit{Optimization}: candidates are sub-sampled and assembled into a sandwich-like model, thus jointly optimized in the \textit{sandwich framework}. \textit{Selection}: the candidate with the best \textit{$\lambda$-tradeoff} is selected, thus the student is distilled in one trail.

\subsection{Scale-performance Tradeoff}

While the scale-performance tradeoff can be an indicator of a good teacher assistant, it is not easy to measure.
To empirically quantify the scale-performance balance, we introduce a new tradeoff measure below:
\begin{definition}[\textit{$\lambda$-tradeoff}] The $\lambda$-tradeoff measure of a teacher assistant $(\mathcal{A}, s_a, m_a)$ is defined as $t_a=m_a+\lambda\cdot (1-s_a)$, where $\lambda \in [0, 1]$.
\end{definition}

In practice, we observe that the \textit{$\lambda$-tradeoff} (\textcolor[rgb]{0.89,0.10,0.11}{red} curves) of the teacher assistant is positively correlated with the performance of the student (\textcolor[rgb]{0.2,0.63,0.17}{green} curves).
Theoretically, due to the linear property of the \textit{$\lambda$-tradeoff} and the concave property of the teacher assistant scale-performance correlation, there should always be one and only one maximum value of \textit{$\lambda$-tradeoff}.

\subsection{Sandwich Framework}

The problem can be reformulated as finding an optimal teacher assistant that has the maximum value of \textit{$\lambda$-tradeoff}:  
\begin{equation}
\small
\begin{aligned}
    (\mathcal{A^*}, s_a^*, m_a^*) &= \underset{\mathcal{A},s_a,m_a}{\mathrm{argmax}} \ t_a \\
    &= \underbrace{\underbrace{\underset{s_a} {\mathrm{argmax}}\ \underset{\mathcal{A}}{\mathrm{argmax}}}_{\text{specification}}\underbrace{\underset{m_a}{\mathrm{argmax}}\ t_a}_{\text{optimization}}}_{\text{selection}}
\end{aligned}
\end{equation}

Based on the above reformulation, a sandwich framework can be implemented to solve the problem with three main stages: \textit{specification}, \textit{optimization}, and \textit{selection}. Essentially, during \textit{specification}, a set of teacher assistant candidates are generated of different scales. Then the performance metric of the teacher assistant of each scale is obtained through an efficient \textit{optimization}. These two stages form a feasible region for the above reformulation. Finally, the optimal teacher assistant $\mathcal{A^*}$ is selected with a linear scanning of the feasible region during \textit{selection}. After the discovery of the optimal teacher assistant, the teacher assistant can subsequently be distilled to the expected student.

\paragraph{Specification}

We use gridding and pruning techniques to identify the structure of each candidate.

\textit{Gridding}. Theoretically, one needs to generate candidates at every possible scale to find the optimal solution. However, it is impossible to enumerate all possibilities in a continuous space. Therefore, we discretize the candidate scales into $n$ discrete values, $\{\mathcal{A} = (\mathcal{A}_k,s_{a_k}, m_{a_k}) \ | \ \Delta s_{a} = (s_t - s_s) / n\}$, with equal slicing between the teacher scale and student scale.

\textit{Pruning}. For candidates at various scales, there are still an infinite number of possible structures, e.g., different combinations of width and depth. A number of approaches have been proposed to identify a good structure at a scale, including dynamic search~\citep{HouHSJCL20}, layer dropping~\citep{FanGJ20} and pruning~\citep{MichelLN19}. In this work, we adopt pruning to assign structures $\mathcal{A}_k$ to the candidates due to its known advantages in knowledge distillation~\citep{XiaZC22}. Concretely, following previous work~\citep{MichelLN19}, the pruning starts with the least important parameters based on their importance scores, which are approximated by masking the parameterized structures. The technical details of our pruning are supplied in Appendix~\ref{app:1}.

Essentially, gridding positions the scales of candidates between the scales of the teacher and student with equal intervals and pruning assigns candidates with pruned structures.

\paragraph{Optimization}

A straightforward solution to unearth the optimality of each candidate is exhaustively measuring the student performance distilled from each, e.g., \textsc{MaxiDisc}. \textit{$\lambda$-tradeoff} offers a chance to measure the optimality without actual distillation. However, the memory footprints and computational costs apparently can also be extremely large considering the number of candidates when obtaining performance (i.e., $m_a$) of all candidates. To reduce the memory overhead and the computational complexity, we introduce two effective approximations, \textit{parameter-sharing} and \textit{sandwich-optimization}, so that the \textit{$\lambda$-tradeoff}s of all candidates at different scales can be yielded in one run.
The feasibility of the approximations are guarded by the following two properties.
\begin{property}[Incremental Property]
For two candidates $\mathcal{A}_i$ and $\mathcal{A}_j$ in the teacher assistant candidate set $\mathcal{A}$, if $s_i < s_j$, then we have $\mathcal{A}_i\subset \mathcal{A}_j$.
\end{property}
This incremental property is an outcome of the pruning approach~\citep{LiKDSG17,FrankleC19}, which essentially tells that among all candidates obtained from the specification, the structure of a candidate at a smaller scale is a subset of the structure for a candidate at a larger scale. 
\begin{remark}
The incremental property affirms that a larger candidate can result in a smaller one by continuously pruning less significant parameters, which enables these candidates to be assembled into one sandwich-like model in a \underline{parameter-sharing} fashion. The memory scale of the sandwich-like model is exactly that of the largest candidate.
\end{remark}

\begin{property}[Sandwich Rule]
For two candidates $\mathcal{A}_i$ and $\mathcal{A}_j$ from candidate set $\mathcal{A}$, if $s_i < s_j$, then we have $m_s\le m_i\le m_j\le m_t$.
\end{property}

The sandwich rule~\citep{YuH19,CaiGWZH20} states that the performance of a candidate is bounded by the best performance of a larger candidate and a smaller one, due to the subset structure. 
Therefore, a candidate can be optimized by alternatively distilling its larger and smaller candidates, without direct distillation.

\begin{remark}
The sandwich rule allows us to sub-sample $\eta$ out of all $n$ ($\eta\le n$) filling-like candidates and conduct \underline{sandwich-optimization} over the sampled candidates, which substantially reduces the computational cost.
\end{remark}

With the two approximations, we reduce the memory footprints of all candidates to a distinguished one via parameter-sharing. The computational costs are also largely reduced with sandwich-optimization.
Finally, we formulate the distillation objectives for task-specific distillation (TSD) and task-agnostic distillation (TAD) respectively as:
\begin{equation}
\small
\begin{aligned}
    \mathcal{L}_{\sf TSD}=&\sum_{i=1}^{\eta}\text{CE}(\mathbf{y}_{\mathcal{T}},\mathbf{y}_{\mathcal{A}_i}) + \text{MSE} (\mathbf{H}_{\mathcal{T}},\mathbf{H}_{\mathcal{A}_i}) \\
    \mathcal{L}_{\sf TAD}=&\sum_{i=1}^{\eta}\text{KL}(\mathbf{R}_{\mathcal{T}}^{\sf Q},\mathbf{R}_{\mathcal{A}_i}^{\sf Q})+\text{KL}(\mathbf{R}_{\mathcal{T}}^{\sf K},\mathbf{R}_{\mathcal{A}_i}^{\sf K})\\
    &+\text{KL}(\mathbf{R}_{\mathcal{T}}^{\sf V},\mathbf{R}_{\mathcal{A}_i}^{\sf V})
\end{aligned}
\end{equation}
where MSE, CE and KL stand for mean squared error, cross entropy and kullback-leibler divergence respectively.
$\mathbf{H}$ is the last layer of hidden states, $\mathbf{y}$ is the final prediction. As is taken from MiniLM~\citep{WangBHDW21}, $\mathbf{R}^{\sf Q}$ is the query relation matrix containing totally $h$ attention heads from the last layer, likewise $\mathbf{R}^{\sf K}$ and $\mathbf{R}^{\sf V}$ are the key and value relation matrices. Since heads can be pruned for a teacher assistant candidate, an additional self-attention module is employed as the last layer for TAD. The teacher assistants with the best performance at different scales can be obtained after the above optimization.
The unsampled teacher assistants can be retrieved based on the larger teacher assistant from the sampled pool using the shared parameters.

\paragraph{Selection}

The optimal teacher assistant can be identified by selecting the candidate with the best \textit{$\lambda$-tradeoff} measure, which is then distilled to the expected student again following above distillation objectives.
Note that the tradeoff measure is also dependent on $\lambda$. However, we empirically find that the optimal solution of \textsc{MiniDisc} is relatively stable with a wide range of $\lambda$, and we fix $\lambda$ to 0.2 in all our experiments. More discussion on the impact of $\lambda$ is provided in the experiments.

\section{Experiments}

\subsection{Setup}

\paragraph{Datasets and Metrics}

We conduct experiments on GLUE~\citep{WangSMHLB19}. The GLUE originally consists of two sequence classification tasks, SST-2~\citep{SocherPWCMNP13} and CoLA~\citep{WarstadtSB19}, with seven sequence-pair classification tasks, i.e., MRPC~\citep{DolanB05}, STS-B~\citep{CerDALS17}, QQP, MNLI~\citep{WilliamsNB18}, QNLI~\citep{RajpurkarZLL16}, RTE~\citep{BentivogliMDDG09} and WNLI~\citep{LevesqueDM12}. We exclude WNLI and CoLA due to the evaluation inconsistency (in other words, compressed LMs get dramatically worse results while original LMs get much better ones as found out in~\citep{XiaZC22}) and use the other seven tasks for evaluation. Following the work in BERT~\citep{DevlinCLT19}, we report F1 on MRPC and QQP, Spearman Correlation scores (Sp Corr) on STS-B, and Accuracy (Acc) on other tasks. Macro average scores (Average) over these seven tasks are computed for overall performance. Results on development sets are reported. We also adopt Wikipedia for pretraining in task-agnostic distillation. The detailed statistics, maximum sequence lengths, and metrics of GLUE and Wikipeida are supplied in Appendix~\ref{app:2}.

\paragraph{Implementation Details}

Experiments are carried out on BERT\textsubscript{\sf base}~\citep{DevlinCLT19} and EncT5\textsubscript{\sf xl}~\citep{LiuSY21}. EncT5 is a language model which achieves competitive performance as T5~\citep{RaffelSRLNMZLL20} on GLUE with a nearly encoder-only T5 (incorporated with a decoder layer). Our task-specific experiments are carried out on either one Nvidia A100 for EncT5\textsubscript{\sf xl} or one Nvidia V100 for BERT\textsubscript{\sf base}, and $\eta$ is set to 6 according to our empirical investigation. On the other hand, the task-agnostic experiments are carried out on eight Nvidia A100s with BERT\textsubscript{\sf base}. $\eta$ is set to 3 to substantially reduce computational burden. The number of relation heads is set to 32 since we use deep relation distillation as the task-agnostic distillation objective. Other implementation details are supplied in Appendix~\ref{app:3}. Generally, the sampling is performed from candidates at scales \{100\%, 95\%, 90\%, $\dots$, 10\%, 5\%\}.

\begin{table*}[ht]
    \caption{The results of task-specific distillation upon BERT\textsubscript{\sf base}. The GPU hours of teacher assistant-based methods are estimated with respect to their conventional counterparts.}
    \begin{adjustbox}{width=0.95\textwidth,center}
    \begin{tabular}{lrcccccccc|r}
    \toprule
      Method & FLOPs & SST-2 & MRPC & STS-B & QQP & MNLI-m/mm & QNLI & RTE & Average & GPUs \\
    \midrule
      BERT\textsubscript{\sf base} & 10.9G & 93.8 & 91.5 & 87.1 & 88.4 & 84.9/84.9 & 91.9 & 71.5 & 86.7 & $-$ \\
    \midrule
      % BERT\textsubscript{\sf 4L}-KD~\citeyearpar{HintonVD15} & 3.6G & 89.6 & 86.9 & 86.4 & 86.1 & 77.7/77.7 & 85.1 & 65.3 & 81.9 & 1$\times$ \\
      % BERT\textsubscript{\sf 4L}-PKD~\citeyearpar{SunCGL19} & 3.6G & 89.9 & 87.6 & 86.4 & 86.0 & 77.7/77.7 & 85.0 & 65.3 & 82.0 & 1$\times$ \\
      % BERT\textsubscript{\sf 4L}-CKD~\citeyearpar{ParkKY21} & 3.6G & 89.6 & 87.2 & 86.4 & 86.2 & 77.7/77.9 & 85.0 & 64.6 & 81.8 & 1$\times$ \\
      % DynaBERT\textsubscript{\sf 30\%}~\citeyearpar{HouHSJCL20} & 3.3G & 90.3 & 87.4 & 87.2 & 86.6 & 81.5/81.8 & 89.1 & 66.1 & 83.7 & 1$\times$ \\
      % BERT\textsubscript{\sf 30\%}-FT~\citeyearpar{LiKDSG17} & 3.3G & 91.9 & 88.5 & 87.2 & 87.7 & 82.0/82.6 & 89.5 & 69.0 & 84.8 & 1$\times$ \\
      % BERT\textsubscript{\sf 30\%}-KD~\citeyearpar{HintonVD15} & 3.3G & 92.0 & 88.9 & 86.8 & 87.8 & 82.2/82.7 & 89.9 & 68.2 & 84.8 & 1$\times$ \\
      % BERT\textsubscript{\sf 30\%}-$\mathcal{L}_{\sf TSD}$ & 3.3G & 91.9 & 89.5 & 86.4 & 88.0 & 82.5/82.8 & 89.9 & 68.6 & \textbf{84.9} & 1$\times$ \\
      \multicolumn{11}{c}{\textit{Conventional Distillation}} \\
    \midrule
      KD\textsubscript{\sf 2L}~\citeyearpar{HintonVD15} & 1.8G & 86.8 & 82.5 & 46.8 & 83.7 & 73.5/73.1 & 79.6 & 58.1 & 73.0 & 1$\times$ \\
      PKD\textsubscript{\sf 2L}~\citeyearpar{SunCGL19} & 1.8G & 86.7 & 82.4 & 46.8 & 83.7 & 73.4/73.0 & 79.7 & 57.4 & 72.9 & 1$\times$ \\
      CKD\textsubscript{\sf 2L}~\citeyearpar{ParkKY21} & 1.8G & 86.4 & 82.3 & 48.6 & 83.6 & 73.3/73.0 & 79.1 & 56.7 & 72.9 & 1$\times$ \\
      StarK\textsubscript{\sf 2L}~\citeyearpar{Yang22} & 1.8G & 88.1 & 83.1 & 48.6 & 83.8 & 73.9/74.3 & 80.4 & 57.8 & 73.7 & 1$\times$ \\
      DynaBERT\textsubscript{\sf 15\%}~\citeyearpar{HouHSJCL20} & 2.2G & 89.1 & 85.1 & 84.7 & 84.3 & 78.3/79.0 & 86.6 & 61.4 & 81.1 & 1$\times$ \\
      FT\textsubscript{\sf 15\%}~\citeyearpar{LiKDSG17} & 1.6G & 89.9 & 87.1 & 85.6 & 86.1 & 79.9/80.1 & 85.7 & 63.9 & 82.3 & 1$\times$ \\
      KD\textsubscript{\sf 15\%}~\citeyearpar{HintonVD15} & 1.6G & 89.9 & 88.6 & 85.1 & 86.2 & 79.8/80.2 & 85.6 & 63.9 & 82.4 & 1$\times$ \\
      $\mathcal{L}_{\sf TSD}\textsubscript{\sf 15\%}$ & 1.6G & 90.1 & 88.9 & 85.1 & 86.5 & 80.0/80.2 & 86.0 & 65.3 & \textbf{82.8} & 1$\times$ \\
      
    \midrule
      FT\textsubscript{\sf 10\%}~\citeyearpar{LiKDSG17} & 1.1G & 88.2 & 84.8 & 84.7 & 84.4 & 77.6/77.3 & 84.3 & 65.3 & 80.8 & 1$\times$ \\
      KD\textsubscript{\sf 10\%}~\citeyearpar{HintonVD15} & 1.1G & 88.2 & 87.6 & 84.0 & 84.4 & 77.6/77.4 & 84.3 & 67.2 & 81.3 & 1$\times$ \\
      $\mathcal{L}_{\sf TSD}$\textsubscript{\sf 10\%} & 1.1G & 88.8 & 87.8 & 84.0 & 84.6 & 77.6/77.5 & 84.9 & 66.4 & \textbf{81.5} & 1$\times$ \\
    \midrule
      FT\textsubscript{\sf 5\%}~\citeyearpar{LiKDSG17} & 0.5G & 85.4 & 82.8 & 84.1 & 82.6 & 72.5/73.3 & 81.7 & 63.9 & 78.3 & 1$\times$ \\
      KD\textsubscript{\sf 5\%}~\citeyearpar{HintonVD15} & 0.5G & 85.6 & 84.0 & 83.8 & 82.5 & 72.6/73.2 & 81.6 & 63.2 & 78.3 & 1$\times$ \\
      $\mathcal{L}_{\sf TSD}$\textsubscript{\sf 5\%} & 0.5G & 85.4 & 85.5 & 83.9 & 82.7 & 73.0/73.4 & 82.7 & 63.2 & \textbf{78.7} & 1$\times$ \\
    \midrule
      \multicolumn{11}{c}{\textit{Teacher Assistant-based Distillation}} \\
    \midrule
        TA\textsubscript{\sf 15\%}~\citeyearpar{MirzadehFLLMG20} & 1.6G & 89.3 & 87.7 & 85.3 & 85.7 & 80.0/80.3 & 88.1 & 68.4 & 83.1 & 2$\times$ \\
        \textsc{MaxiDisc}\textsubscript{\sf 15\%} & 1.6G & 89.8 & 87.7 & 85.4 & 86.9 & 81.0/80.1 & 86.1 & 68.2 & 83.2 & 40$\times$ \\
      \rowcolor{orange!20} \textsc{MiniDisc}\textsubscript{\sf 15\%} & 1.6G & 89.8 & 88.2 & 85.8 & 86.6 & 80.3/79.9 & 87.3 & 68.2 & \textbf{83.3} & 4$\times$ \\
    \midrule
      TA\textsubscript{\sf 10\%}~\citeyearpar{MirzadehFLLMG20} & 1.1G & 89.1 & 87.9 & 83.1 & 84.7 & 77.8/77.9 & 85.7 & 68.6 & 81.8 & 2$\times$ \\
      \textsc{MaxiDisc}\textsubscript{\sf 10\%} & 1.1G & 89.0 & 88.2 & 84.8 & 84.8 & 78.3/77.8 & 85.3 & 66.8 & 81.9 & 40$\times$ \\
      \rowcolor{orange!20} \textsc{MiniDisc}\textsubscript{\sf 10\%} & 1.1G & 89.1 & 88.4 & 85.4 & 84.9 & 78.2/78.6 & 86.3 & 68.2 & \textbf{82.4} & 4$\times$ \\
    \midrule
      TA\textsubscript{\sf 5\%}~\citeyearpar{MirzadehFLLMG20} & 0.5G & 86.5 & 86.5 & 82.2 & 83.2 & 73.3/73.7 & 82.6 & 65.3 & 79.2 & 2$\times$ \\
      \textsc{MaxiDisc}\textsubscript{\sf 5\%} & 0.5G & 86.9 & 88.3 & 84.8 & 83.7 & 74.4/76.3 & 83.5 & 65.0 & \textbf{80.4} & 40$\times$ \\
      \rowcolor{orange!20} \textsc{MiniDisc}\textsubscript{\sf 5\%} & 0.5G & 86.9 & 87.6 & 84.8 & 83.5 & 72.7/74.5 & 84.0 & 66.8 & 80.1 & 4$\times$ \\
    \bottomrule
    \end{tabular}
    \end{adjustbox}
    \label{tab:1}
    \vspace{-3mm}
\end{table*}

\paragraph{Baselines}

We compare our model with several state-of-the-art baselines. \textsubscript{\sf *L;*H} denotes dropping layers and hidden dimensions, while \textsubscript{\sf *\%} represents structured pruning with either local ranking or our global ranking. 
%(see Appendix~\ref{app:1}).
\begin{itemize}
    \item \textbf{Conventional Distillation:} FT~\citep{LiKDSG17} indicates direct finetuning after pruning.
    KD~\citep{HintonVD15}, PKD~\citep{SunCGL19} and CKD~\citep{ParkKY21} are methods with different objectives, i.e., KD directly distills logits, PKD distills both logits and hidden states and CKD distills token and layer relations. DynaBERT~\citep{HouHSJCL20} uses structured pruning with a local ranking in each layer. StarK~\citep{Yang22} views sparse teachers as student-friendly teachers. MiniLM~\citep{WangBHDW21} is distilled with the deep relation alignment. TinyBERT~\citep{JiaoYSJCL0L20} is distilled with a combination of various feature distillations.
    \item \textbf{Teacher Assistant-based Distillation:} 
    TA~\citep{MirzadehFLLMG20,WangW0B0020} is specifically incorporated for both task-specific and task-agnostic distillation with a 40\%-scale teacher assistant. \textsc{MaxiDisc} goes further upon TA and manually selects the best teacher assistant among available trials.
\end{itemize}

\begin{table*}[ht]
    \caption{The results of task-specific distillation upon EncT5\textsubscript{\sf xl}. The GPU hours of teacher assistant-based methods are estimated with respect to their conventional counterparts.}
    \begin{adjustbox}{width=0.90\textwidth,center}
    \begin{tabular}{lrcccccccc|r}
    \toprule
      Method & FLOPs & SST-2 & MRPC & STS-B & QQP & MNLI-m/mm & QNLI & RTE & Average & GPUs \\
    \midrule
      EncT5\textsubscript{\sf xl} & 155.9G & 96.9 & 95.1 & 92.3 & 90.0 & 90.7/90.9 & 95.0 & 88.5 & 92.4 & $-$ \\
    \midrule
      \multicolumn{11}{c}{\textit{Conventional Distillation}} \\
    \midrule
      FT\textsubscript{\sf 10\%}~\citeyearpar{LiKDSG17} & 15.6G & 91.6 & 87.1 & 86.7 & 87.9 & 81.9/87.0 & 66.1 & 91.6 & 83.8 & 1$\times$ \\
      KD\textsubscript{\sf 10\%}~\citeyearpar{HintonVD15} & 15.6G & 92.2 & 86.8 & 86.6 & 87.9 & 83.6/83.8 & 88.1 & 63.5 & 84.1 & 1$\times$ \\
      $\mathcal{L}_{\sf TSD}$\textsubscript{\sf 10\%} & 15.6G & 94.5 & 90.2 & 87.4 & 87.9 & 84.7/84.1 & 90.8 & 67.5 & \textbf{85.9} & 1$\times$ \\
    \midrule
      FT\textsubscript{\sf 5\%}~\citeyearpar{LiKDSG17} & 7.8G & 90.1 & 84.8 & 84.7 & 86.5 & 78.0/78.2 & 83.9 & 62.8 & 81.1 & 1$\times$ \\
      KD\textsubscript{\sf 5\%}~\citeyearpar{HintonVD15} & 7.8G & 89.9 & 85.1 & 85.4 & 86.6 & 79.4/79.6 & 84.2 & 55.6 & 80.7 & 1$\times$ \\
      $\mathcal{L}_{\sf TSD}$\textsubscript{\sf 5\%} & 7.8G & 92.9 & 88.0 & 83.4 & 85.4 & 79.6/80.0 & 87.0 & 58.8 & \textbf{81.9} & 1$\times$ \\
    \midrule
      \multicolumn{11}{c}{\textit{Teacher Assistant-based Distillation}} \\
    \midrule
      TA\textsubscript{\sf 10\%} & 15.6G & 94.5 & 90.7 & 87.4 & 88.0 & 85.2/84.6 & 91.1 & 69.3 & 86.3 & 2$\times$ \\
      \textsc{MaxiDisc}\textsubscript{\sf 10\%} & 15.6G & 94.6 & 90.5 & 88.0 & 88.1 & 86.2/85.1 & 91.5 & 70.4 & 86.8 & 40$\times$ \\
      \rowcolor{orange!20} \textsc{MiniDisc}\textsubscript{\sf 10\%} & 15.6G & 94.6 & 91.5 & 87.8 & 87.3 & 85.9/85.0 & 91.1 & 72.2 & \textbf{86.9} & 4$\times$ \\
    \midrule
      TA\textsubscript{\sf 10\%} & 7.8G & 92.3 & 88.4 & 83.7 & 86.0 & 80.2/80.5 & 87.5 & 56.3 & 81.9 & 2$\times$ \\
      \textsc{MaxiDisc}\textsubscript{\sf 10\%} & 7.8G & 93.0 & 88.0 & 83.9 & 86.5 & 81.2/81.6 & 88.1 & 67.5 & 83.7 & 40$\times$ \\
      \rowcolor{orange!20} \textsc{MiniDisc}\textsubscript{\sf 10\%} & 7.8G & 93.8 & 89.8 & 85.3 & 86.7 & 82.9/82.7 & 89.2 & 64.6 & \textbf{84.4} & 4$\times$ \\
    \bottomrule
    \end{tabular}
    \end{adjustbox}
    \label{tab:2}
    \vspace{-3mm}
\end{table*}

\begin{table}[ht]
    \centering
    \caption{The results of task-specific distillation upon LLaMA2\textsubscript{\sf 7B}. The Alpaca dataset~\citep{alpaca} is utilized as the distillation data.}
    \begin{adjustbox}{width=0.22\textwidth,center}
    \begin{tabular}{lc}
    \toprule
        Method & MMLU \\
    \midrule
        LLaMA2\textsubscript{\sf 7B} & 46.0 \\
    \midrule
        KD\textsubscript{\sf 15\%} & 25.6 \\
    \midrule
        TA\textsubscript{\sf 15\%} & 26.1 \\
        \textsc{MaxiDisc}\textsubscript{\sf 15\%} & 26.8 \\
        \textsc{MiniDisc}\textsubscript{\sf 15\%} & 26.9 \\
    \bottomrule
    \end{tabular}
    \end{adjustbox}
    \label{tab:a}
\end{table}

\begin{table*}[ht]
    \caption{The results of task-agnostic distillation upon BERT\textsubscript{\sf base}. The results of TinyBERT are reproduced based on their released checkpoints without additional task-specific distillation for a fair comparison. The GPU hours of teacher assistant-based methods are estimated with respect to their conventional counterparts.}
    \begin{adjustbox}{width=0.95\textwidth,center}
    \begin{tabular}{lrcccccccc|r}
    \toprule
      Method & FLOPs & SST-2 & MRPC & STS-B & QQP & MNLI-m/mm & QNLI & RTE & Average & GPUs \\
    \midrule
      BERT\textsubscript{\sf base} & 10.9G & 93.8 & 91.5 & 87.1 & 88.4 & 84.9/84.9 & 91.9 & 71.5 & 86.7 & $-$ \\
    \midrule
      \multicolumn{11}{c}{\textit{Conventional Distillation}} \\
    \midrule
      FT\textsubscript{\sf 10\%}~\citeyearpar{LiKDSG17} & 1.1G & 84.6 & 83.1 & 83.8 & 84.5 & 75.3/75.4 & 83.2 & 56.7 & 78.3 & 1$\times$ \\
      $\mathcal{L}_{\sf TSD}$\textsubscript{\sf 10\%} & 1.1G & 90.7 & 89.0 & 87.0 & 85.9 & 78.4/78.2 & 86.0 & 66.4 & 82.7 & 1$\times$ \\
      MiniLM\textsubscript{\sf 4L;384H}~\citeyearpar{WangBHDW21} & 0.9G & 90.0 & 88.6 & 87.2 & 86.1 & 80.0/80.3 & 87.9 & 67.2 & 83.4 & 1$\times$ \\
      $\mathcal{L}_{\sf TAD}$\textsubscript{\sf 10\%} & 1.1G & 92.0 & 90.1 & 87.9 & 86.6 & 80.0/80.3 & 88.0 & 67.2 & \textbf{84.0} & 1$\times$ \\
    \midrule
      FT\textsubscript{\sf 5\%}~\citeyearpar{LiKDSG17} & 0.5G & 84.1 & 82.4 & 81.8 & 83.7 & 74.4/74.9 & 82.5 & 57.0 & 77.6 & 1$\times$ \\
      TinyBERT\textsubscript{\sf 4L;312H}~\citeyearpar{JiaoYSJCL0L20} & 0.6G & 88.5 & 87.9 & 86.6 & 85.6 & 78.9/79.2 & 87.3 & 67.2 & 82.7 & 1$\times$ \\
      MiniLM\textsubscript{\sf 3L;384H}~\citeyearpar{WangBHDW21} & 0.7G & 89.1 & 89.1 & 86.6 & 85.4 & 77.8/78.4 & 87.2 & 66.1 & 82.5 & 1$\times$ \\
      $\mathcal{L}_{\sf TAD}$\textsubscript{\sf 5\%} & 0.5G & 90.9 & 89.4 & 87.7 & 85.8 & 79.2/79.8 & 87.3 & 65.7 & \textbf{83.2} & 1$\times$ \\
    \midrule
      \multicolumn{11}{c}{\textit{Teacher Assistant-based Distillation}} \\
    \midrule
      TA\textsubscript{\sf 10\%}~\citeyearpar{WangW0B0020} & 0.9G & 90.0 & 88.5 & 87.3 & 86.3 & 80.1/80.7 & 88.0 & 66.4 & 83.4 & 2$\times$ \\
      \textsc{MaxiDisc}\textsubscript{\sf 10\%} & 1.1G & 91.5 & 90.3 & 87.8 & 86.6 & 80.0/80.1 & 88.6 & 67.2 & \textbf{84.0} & 40$\times$ \\
      \rowcolor{orange!20} \textsc{MiniDisc}\textsubscript{\sf 10\%} & 1.1G & 91.4 & 90.0 & 87.5 & 86.6 & 79.8/80.0 & 88.0 & 67.2 & 83.8 & 4$\times$ \\
    \midrule
      TA\textsubscript{\sf 5\%}~\citeyearpar{WangW0B0020} & 0.7G & 89.8 & 85.9 & 86.0 & 85.5 & 77.6/78.5 & 86.8 & 66.1 & 82.0 & 2$\times$ \\
      \textsc{MaxiDisc}\textsubscript{\sf 5\%} & 0.5G & 90.1 & 89.7 & 87.4 & 85.6 & 79.3/79.7 & 87.1 & 67.9 & \textbf{83.4} & 40$\times$ \\
      \rowcolor{orange!20} \textsc{MiniDisc}\textsubscript{\sf 5\%} & 0.5G & 89.3 & 89.7 & 87.4 & 85.9 & 79.2/79.4 & 86.9 & 69.7 & \textbf{83.4} & 4$\times$ \\
    \bottomrule
    \end{tabular}
    \end{adjustbox}
    \label{tab:3}
\end{table*}

\subsection{Main Results}

\paragraph{Results of Task-specific Distillation}

Table~\ref{tab:1} presents the comparison results of different methods on task-specific distillation at three student scales. There are several key observations: \textbf{First}, both \textsc{MiniDisc} and \textsc{MaxiDisc} yield better performance than TA does and \textsc{MiniDisc} obtains similar or even better results compared to \textsc{MaxiDisc} with much fewer GPU hours. This validates the efficiency of \textsc{MiniDisc} for identifying a good teacher assistant. Notably, the slight performance improvement is attributed to parameter sharing, which is detailed in later analysis. For further smaller BERT\textsubscript{\sf 3\%}, the result still holds, as supplied in Appendix~\ref{app:4}. Additional comparisons of practical inference measurement are supplied in Appendix~\ref{app:5}. \textbf{Second}, pruning based models perform much better compared to the layer dropping methods, e.g., KD\textsubscript{\sf 15\%} achieves much higher score than FLOPs-matched KD\textsubscript{\sf 2L}, which verifies the effectiveness of pruning approach in knowledge distillation. Moreover, we discover the global ranking strategy surpasses the local ranking one by comparing $\mathcal{L}_{\sf TSD}$\textsubscript{\sf 15\%} to FLOPs-matched DynaBERT\textsubscript{\sf 15\%}. We speculate the structures induced by the local ranking strategy are not that effective. The distribution of example pruned structures is supplied in Appendix~\ref{app:6}. \textbf{Third}, conventional distillation methods generate reasonable results at large student scale but fail to maintain the student performance at small scale. Nonetheless, TA consistently outperforms the conventional baselines at all scales. 

\paragraph{Results of Large-scale Distillation}

As is shown in Table~\ref{tab:2}, we conduct a similar comparison on a large LM, EncT5\textsubscript{\sf xl}, with over one billion parameters. The very first results of the large LM also exhibit an akin trend as the one in BERT\textsubscript{\sf base}. The results on a more recent large LM LLaMA2\textsubscript{\sf 7B} are displayed in Table~\ref{tab:a}. And the results on a moderate BERT\textsubscript{\sf large} are supplied in Appendix~\ref{app:7}. We therefore conclude that the scalability of \textsc{MiniDisc} is also compelling. Reversely, the results of \textsc{MiniDisc} on small LMs are supplied in Appendix~\ref{app:8}.

\paragraph{Results of Task-agnostic Distillation}

We also apply \textsc{MiniDisc} to task-agnostic distillation and report the results in Table~\ref{tab:3}. 
The first glimpse is that $\mathcal{L}_{\sf TAD}$ surpasses $\mathcal{L}_{\sf TSD}$, indicating the deep relation alignment is more suitable for task-agnostic distillation. Surprisingly, we discover that the pruned structures can boost the performance of MiniLM, i.e., $\mathcal{L}_{\sf TAD}$, and establish a new state-of-the-art for conventional task-agnostic distillation. 
Another interesting observation is that teacher assistant-based distillation methods do not improve the performance over conventional distillation methods until the scale is reduced to 5\%, indicating that conventional distillation methods are already promising choices on task-agnostic distillation at large scales.
Nonetheless, we still argue the applicability of \textsc{MiniDisc} to task-agnostic distillation for a performance guarantee. 
Note that the results of TinyBERT with additional task-specific distillation are supplied in Appendix~\ref{app:9}.

\subsection{Analyses}

\begin{figure}[ht]
    \centering
    \includegraphics[width=0.47\textwidth]{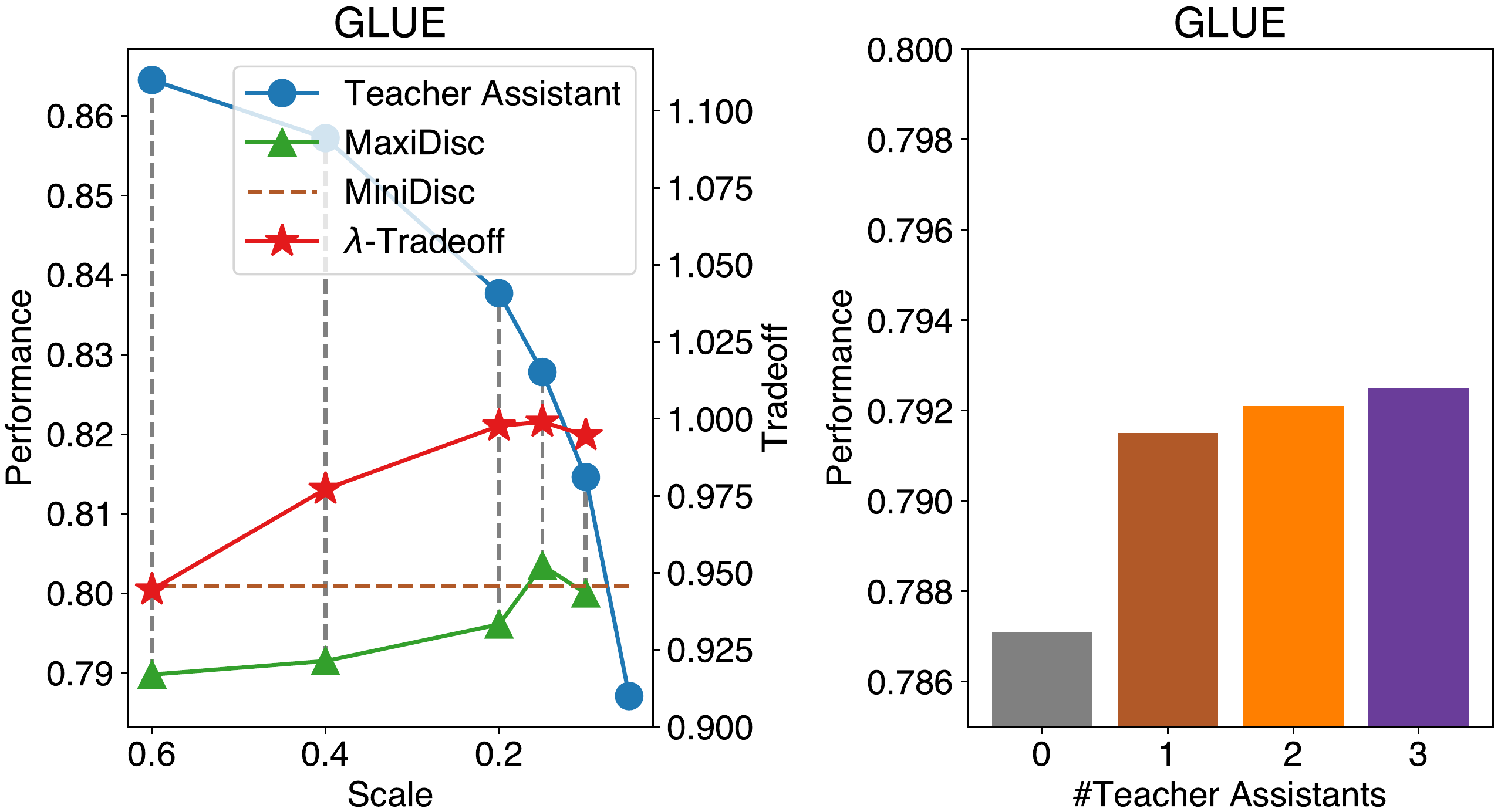}
    \caption{Tradeoff studies by distilling the teacher to a student at 5\% scale. On the left hand, the \textcolor[rgb]{0.12,0.47,0.71}{blue} curve represents the performance of teacher assistants at different scales. The \textcolor[rgb]{0.2,0.63,0.17}{green} curve represents the performance of \textsc{MaxiDisc} using these teacher assistants. The \textcolor[rgb]{0.89,0.10,0.11}{red} curve represents the \textit{$\lambda$-tradeoff} value. The \textcolor[rgb]{0.69,0.35,0.16}{brown} dashed line represents the performance of \textsc{MiniDisc}. On the right hand, the \textcolor[rgb]{0.69,0.35,0.16}{brown}, \textcolor[rgb]{1.0,0.50,0.0}{orange}, and \textcolor[rgb]{0.42,0.24,0.60}{purple} bars represent the performance of \textsc{MiniDisc} using one, two, and three teacher assistants.}
    \label{fig:3}
    \vspace{-3mm}
\end{figure}

\begin{table}[ht]
    \centering
      \caption{The ablation study upon distilling BERT\textsubscript{\sf base} to BERT\textsubscript{\sf 10\%}.}
      \begin{adjustbox}{width=0.49\textwidth,center}
      \begin{tabular}{lrcc}
      \toprule
        Method & GPU hours & MRPC & QQP \\
      \midrule
        $\mathcal{L}_{\sf TSD}$\textsubscript{\sf 10\%} & 1$\times$ & 87.8 & 84.6 \\
        \textsc{MaxiDisc}\textsubscript{\sf 10\%} & 40$\times$ & 88.2 & 84.8 \\
        \qquad w/ \textit{$\lambda$-tradeoff} & 21$\times$ & 88.2 & 84.8 \\
        \qquad w/ \textit{sandwich framework} & 23$\times$ & 88.4 & 84.9 \\
        \textsc{MiniDisc}\textsubscript{\sf 10\%} & 4$\times$ & 88.4 & 84.9 \\
      \bottomrule
      \end{tabular}
      \end{adjustbox}
      \label{tab:4}
      \vspace{-3mm}
\end{table}

\paragraph{Ablation Study}

We carry out an ablation study can actually be viewed as a process of bridging \textsc{MaxiDisc} to \textsc{MiniDisc} by firstly adding \textit{$\lambda$-tradeoff}, then adding \textit{sandwich framework}. We present the results in Table~\ref{tab:4}. The results show that: 1) (\textsc{MaxiDisc} v.s. \textsc{MaxiDisc} w/ \textit{$\lambda$-tradeoff}) \textit{$\lambda$-tradeoff} can be an accurate measure to select the optimal teacher assistant; 2) (\textsc{MaxiDisc} v.s. \textsc{MaxiDisc} w/ \textit{sandwich framework}) \textit{sandwich framework} can achieve competitive (even slightly better) performance despite the \textit{parameter sharing} among teacher assistant candidates; 3) (\textsc{MaxiDisc} w/ \textit{sandwich framework} v.s. \textsc{MiniDisc}) the two together lead to results slightly better than those of \textsc{MaxiDisc} in a much more efficient manner.

\begin{table}[ht]
    \centering
      \caption{The impact of candidate sampling upon distilling BERT\textsubscript{\sf base} to BERT\textsubscript{\sf 10\%}.}
      \begin{adjustbox}{width=0.39\textwidth,center}
      \begin{tabular}{lrc}
      \toprule
        Method & GPU hours & Average \\
      \midrule
        $\mathcal{L}_{\sf TSD}$\textsubscript{\sf 10\%} & 1$\times$ & 81.5 \\
        \textsc{MaxiDisc}\textsubscript{\sf 10\%} & 40$\times$ & 81.9 \\
        \textsc{MiniDisc}\textsubscript{\sf 10\%} ($\eta$=1) & 2$\times$ & 82.1 \\
        \textsc{MiniDisc}\textsubscript{\sf 10\%} ($\eta$=3) & 2$\times$ & 81.9 \\
        \textsc{MiniDisc}\textsubscript{\sf 10\%} ($\eta$=6) & 4$\times$ & 82.4 \\
        \textsc{MiniDisc}\textsubscript{\sf 10\%} ($\eta$=9) & 4$\times$ & 82.4 \\
      \bottomrule
      \end{tabular}
      \end{adjustbox}
      \label{tab:5}
\vspace{-3mm}
\end{table}

\paragraph{Impact of Candidate Sampling}

We then study the impact of the \textit{sandwich framework} in \textsc{MiniDisc} by varying the number of sampled candidates $\eta$, and measuring the training cost and the student performance. From Table~\ref{tab:5}, we show the assembled sandwich together with sub-sampled fillings brings acceptable performance detriment and efficiency gain.

\begin{table}[ht]
    \centering
      \caption{The impact of $\lambda$ upon distilling BERT\textsubscript{\sf base} to BERT\textsubscript{\sf 10\%}.}
      \begin{adjustbox}{width=0.34\textwidth,center}
      \begin{tabular}{lcc}
      \toprule
        Method & MRPC & QQP \\
        \midrule
        $\mathcal{L}_{\sf TSD}$\textsubscript{\sf 10\%} & 87.8 & 84.6 \\
        \textsc{MaxiDisc}\textsubscript{\sf 10\%} & 88.2 & 84.8 \\
        \textsc{MiniDisc}\textsubscript{\sf 10\%} ($\lambda$=0.1) & 87.5 & 85.2 \\
        \textsc{MiniDisc}\textsubscript{\sf 10\%} ($\lambda$=0.2) & 88.4 & 84.9 \\
        \textsc{MiniDisc}\textsubscript{\sf 10\%} ($\lambda$=0.3) & 87.5 & 84.7 \\
        \textsc{MiniDisc}\textsubscript{\sf 10\%} ($\lambda$=0.5) & 87.8 & 84.7 \\
        \textsc{MiniDisc}\textsubscript{\sf 10\%} ($\lambda$=0.7) & 87.8 & 84.7 \\
      \bottomrule
      \end{tabular}
      \end{adjustbox}
      \label{tab:6}
\vspace{-3mm}
\end{table}

\paragraph{Impact of $\lambda$}

To show \textit{$\lambda$-tradeoff} is robust on the value of $\lambda$, we vary $\lambda$ within \{0.1,0.2,0.3,0.5,0.7\}. It can be seen from Table~\ref{tab:6} that the performance of \textsc{MiniDisc} is relatively stable with different values of $\lambda$.
Moreover, we offer a $\lambda$-independent solution using a negative derivative of performance to scale as the tradeoff measure, which yields slightly worse results, as supplied in Appendix~\ref{app:10}. 

\paragraph{Existence of Tradeoff}

To double-check the existence of the concerned tradeoff, we use teacher assistants at different scales within \textsc{MaxiDisc} and plot performance variations of these schedules upon BERT\textsubscript{\sf base} in Figure~\ref{fig:3} (left). It can be seen that reducing the teacher assistant scale can lead to student performance improvement until a certain scale, after which performance degradation is witnessed. All schedules underperform the \textit{$\lambda$-tradeoff} indicated one. 
We attribute the inferiority to improper scale-performance tradeoffs, as concentrating only on either scale or performance will give rise to a trivial solution with pareto optimality~\citep{SenerK18,LinZ0ZK19}. The overall phenomenon implies the existence of scale-performance tradeoff. Similar phenomenon is also observed in EncT5, which is supplied in Appendix~\ref{app:11}.

\paragraph{Sufficiency of One Teacher Assistant}

To examine whether one teacher assistant is sufficient, we insert more than one teacher assistant to \textsc{MiniDisc} and present the results in Figure~\ref{fig:3} (right). It is clear that there is no obvious performance gain when applying more than one teacher assistant (two and three) in schedules. 
Therefore, we alternatively choose to use only one teacher assistant in \textsc{MiniDisc} for training efficiency based on the sufficiency. The conclusion still holds for EncT5, which is supplied in Appendix~\ref{app:11}.

Recently proposed progressive distillation methods~\citep{LiLRLZ021,Lin22}, where students are learned firstly from a small teacher then from a larger teacher, inspire us to inspect whether the same regime could further boost \textsc{MiniDisc} since teacher assistants are essentially small teachers and a natural follow-up action is residually distilling the students from the original teachers (residual distillation). The residual distillation can possibly further improve the performance of \textsc{MiniDisc}, as detailed in Appendix~\ref{app:12}.

% \vspace{-2mm}
\section{Conclusions}

In this paper, we propose \textsc{MiniDisc} to identify an optimal teacher assistant for teacher assistant-based distillation in minimally one trial in contrast to \textsc{MaxiDisc}. 
Having observed that the scale-performance tradeoff of the teacher assistant is of great importance to the performance of the student, we introduce a \textit{$\lambda$-tradeoff} measure that quantifies the scale-performance tradeoff of the teacher assistant, and show that it is positively correlated with the student performance. To efficiently compute the measures for teacher assistant candidates and select the optimal one, we design a sandwich optimization for these candidates. Comprehensive results demonstrate the improved efficiency of \textsc{MiniDisc}.

\section*{Limitations}

Although the value of $\lambda$ is relatively stable in a wide range, the core limitation of \textsc{MiniDisc} is that the value of $\lambda$ should be calibrated before practical use. To enable a more automatic process, we conduct some preliminary study by introducing another metric, which does not require any hyperparameters. More details can be found in Appendix \ref{app:10}. We plan to investigate more along this direction in the future. Another limitation of this work is that we leverage gridding and pruning to identify the model structure of each candidate to ensure these candidate structures satisfying certain property for one-run optimization. However, the gridding and pruning process might yield a sub-optimal model architecture at a given model scale. In future, we also plan to explore how to efficient identify an optimal model structure.

% \section*{Ethics Statement}

% \section*{Acknowledgements}

% Entries for the entire Anthology, followed by custom entries
\bibliography{anthology,custom}

\clearpage
\appendix

\section{Technical Details of Pruning}
\label{app:1}

Concretely, following previous work~\citep{MichelLN19}, the pruning always starts with the least important parameters, which are identified according to importance scores. The importance scores are approximated by first masking the parameterized structures. $\mu_{i}$, $\nu_{i}$, and $\xi_{j}$ denote the mask variables respectively for a self-attention head, optionally a cross-attention head, and a feed-forward neuron, such that for an intermediate input $\mathbf{X}$ and potentially an encoder-produced input $\mathbf{E}$:
\begin{gather}
\begin{aligned}
    &\mathbf{Z}=\text{SelfAttention}(\mathbf{X}) \\
    &=\sum_{i}^{h}\mu_{i}\cdot\text{softmax}(\mathbf{X}\mathbf{W}^{\sf Q}_{i}\mathbf{W}^{\sf K\top}_{i}\mathbf{X}^{\top})\mathbf{X}\mathbf{W}^{\sf V}_{i}\mathbf{W}^{\sf O}_{i},
\end{aligned} \\
\begin{aligned}
    &\mathbf{Z}=\text{CrossAttention}(\mathbf{Z},\mathbf{E}) \\
    &=\sum_{i}^{h}\nu_{i}\cdot\text{softmax}(\mathbf{Z}\mathbf{W}^{\sf Q^{\prime}}_{i}\mathbf{W}^{\sf K^{\prime}\top}_{i}\mathbf{E}^{\top})\mathbf{E}\mathbf{W}^{\sf V^{\prime}}_{i}\mathbf{W}^{\sf O^{\prime}}_{i},
\end{aligned} \\
    \widetilde{\mathbf{X}}=\text{FeedForward}(\mathbf{Z})=\sum_{j}^{d}\xi_{j}\cdot g(\mathbf{Z}\mathbf{W}^{\sf 1}_{j})\mathbf{W}^{\sf 2}_{j},
\end{gather}
where potential bias terms (e.g., linear bias and position bias) are omitted, $i$ means $i$-th head among $h$ heads, $j$ means $j$-th intermediate neuron among $d$ neurons, and $g$ is an activation function. We initialize all mask variables to ones to preserve the original structure at the very beginning.

Then expected absolute gradients over either finetuning or pretraining data gives the important scores:
\begin{gather}
\mathbb{I}_{i}^{\mu}=\mathbb{E}_{(x,y)\sim\mathcal{D}}\left|\frac{\partial\mathcal{L}(x,y)}{\partial\mu_{i}}\right|, \\
\mathbb{I}_{i}^{\nu}=\mathbb{E}_{(x,y)\sim\mathcal{D}}\left|\frac{\partial\mathcal{L}(x,y)}{\partial\nu_{i}}\right|, \\
\mathbb{I}_{j}^{\xi}=\mathbb{E}_{(x,y)\sim\mathcal{D}}\left|\frac{\partial\mathcal{L}(x,y)}{\partial \xi_{j}}\right|,
\end{gather}
where $(x,y)$ is a data point and $\mathcal{L}$ is the task-specific loss for task-specific models or the language modeling loss for pretrained models. $\mathbb{E}$ represents expectation. The absolute value of gradient for a mask indicates how large the impact of pruning the corresponding structure is, thus implying how important the structure is.

Intuitively, we take a global ranking, in contrast to a local one as in other literature~\citep{HouHSJCL20}, for the structures of the same type (i.e., attention head or feed-forward element) from all stacking layers for pruning preference, before which we also normalize the importance scores for same-type structures in a layer with $\ell_2$ norm, as suggested by~\citet{MolchanovTKAK17}, for a balanced pruning. Therefore, for each candidate, we separately prune attention heads and feed-forward elements to the scale so that we reach a qualified structure. For the sake of a corner case that all structures in a module are pruned, we skip the module by feeding the input as the output. While we can alternate to an quite recent pruning method~\citep{XiaZC22} exploiting both coarse-grained and fine-grained strategies for state-of-the-art performance, we argue that our framework is agnostic to pruning methods and keep the pruning method simple.

\section{Dataset Statistics}
\label{app:2}

We conduct experiments on seven datasets. The detailed statistics, maximum sequence lengths, and metrics for datasets we use are shown in Table~\ref{tab:7}, where the Wikipedia corpus used for pretraining is also attached.

\begin{table*}[t]
    \centering
    \caption{The statistics, maximum sequence lengths, and metrics.}
    \begin{adjustbox}{width=0.67\textwidth,center}
    \begin{tabular}{lrrcc}
      \toprule
        Dataset & \#Train exam. & \#Dev exam. & Max. length & Metric \\
      \midrule
        SST-2 &  67K & 0.9K & 64 & Accuracy \\
        MRPC & 3.7K & 0.4K & 128 & F1 \\
        STS-B & 7K & 1.5K & 128 & Spearman Correlation \\
        QQP & 364K & 40K & 128 & F1 \\
        MNLI-m/mm & 393K & 20K & 128 & Accuracy \\
        QNLI & 105K & 5.5K & 128 & Accuracy \\
        RTE & 2.5K & 0.3K & 128 & Accuracy \\
     \midrule
        Wikipedia & 35M & - & 128 & - \\
      \bottomrule
    \end{tabular}
    \end{adjustbox}
    \label{tab:7}
\end{table*}

\section{Additional Implementation Details}
\label{app:3}

The summary of hyperparameters for both task-specific and task-agnostic distillation is shown in Table~\ref{tab:8}.

\begin{table*}[t]
    \centering
    \caption{The hyperparameters for both task-specific and task-agnostic distillation. The learning rate is searched within different grids for BERT\textsubscript{\sf base} and EncT5\textsubscript{\sf xl}.}
    \begin{adjustbox}{width=0.69\textwidth,center}
    \begin{tabular}{lcc}
      \toprule
        Hyperparameter & Task-specific Distillation & Task-agnostic Distillation \\
      \midrule
        Batch Size & \{16,32\} & 8$\times$128$=$1024 \\
        Optimizer & AdamW & AdamW \\
        Learning Rate & \{1e-5, 2e-5, 3e-5\}/\{1e-4, 2e-4, 3e-4\} & 3e-4 \\
        Training Epochs & 10 & 5 \\
        Early-stop Epochs & 5 & - \\
        Warmup Proportion & 0.1 & 0.01 \\
        Weight Decay & 0.01 & 0.01 \\
        Sampling Number $\eta$ & 6 & 3 \\
      \bottomrule
    \end{tabular}
    \end{adjustbox}
    \label{tab:8}
\end{table*}

\section{Additional Results upon \texorpdfstring{BERT\textsubscript{\sf base}}{BERT base}}
\label{app:4}
We further conduct experiments on extremely small scale student model, i.e., BERT\textsubscript{\sf 3\%}. The results are shown in Table~\ref{tab:9}.

\begin{table*}[t]
    \caption{Additional results of task-specific distillation upon BERT\textsubscript{\sf base}.}
    \begin{adjustbox}{width=0.79\textwidth,center}
    \begin{tabular}{lrcccccccc}
    \toprule
      Method & FLOPs & SST-2 & MRPC & STS-B & QQP & MNLI-m/mm & QNLI & RTE & Average \\
    \midrule
      $\mathcal{L}_{\sf TSD}$\textsubscript{\sf 3\%} & 0.3G & 85.2 & 83.6 & 81.9 & 82.1 & 71.9/72.7 & 81.9 & 57.4 & 77.1 \\
      \textsc{MaxiDisc}\textsubscript{\sf 3\%} & 0.3G & 85.6 & 85.0 & 82.7 & 82.7 & 72.7/72.8 & 82.0 & 59.6 & 77.9 \\
      \textsc{MiniDisc}\textsubscript{\sf 3\%} & 0.3G & 85.9 & 85.7 & 83.6 & 83.1 & 72.9/73.6 & 81.9 & 58.1 & 78.1 \\
    \bottomrule
    \end{tabular}
    \end{adjustbox}
    \label{tab:9}
\end{table*}

\section{Practical Inference Measurement}
\label{app:5}

Since FLOPs only offers theoretical inference compute, we additionally provide throughput for empirical inference compute of each model with throughput (i.e., processed tokens per micro second) in Table~\ref{tab:10}. The test environment is established by feeding 32$\times$128 tokens to models. The amount of decomposed parameters is also attached for a reference.

\begin{table*}[ht]
    \centering
    \caption{Inference compute measurement.}
    \begin{adjustbox}{width=0.53\textwidth,center}
    \begin{tabular}{lrrrr}
    \toprule
      Method & FLOPs & Throughput & Trm params & Emb params \\
    \midrule
      BERT\textsubscript{\sf base} & 10.9G & 55.7tokens/ms & 85.7M & 23.8M \\
      BERT\textsubscript{\sf 10\%} & 1.1G & 278.2tokens/ms & 9.1M & 23.8M \\
      BERT\textsubscript{\sf 5\%} & 0.5G & 412.9tokens/ms & 4.9M & 23.8M\\
    \midrule
      BERT\textsubscript{\sf large} & 38.7G & 17.9tokens/ms & 303.3M & 31.8M \\
      BERT\textsubscript{\sf 10\%} & 3.9G & 104.1tokens/ms & 31.3M & 31.8M \\
      BERT\textsubscript{\sf 5\%} & 1.9G & 154.2tokens/ms & 16.3M & 31.8M \\
    \midrule
      EncT5\textsubscript{\sf xl} & 155.8G & 4.8tokens/ms & 1275.1M & 32.9M \\
      EncT5\textsubscript{\sf 10\%} & 15.6G & 38.8tokens/ms & 127.4M & 32.9M \\
      EncT5\textsubscript{\sf 5\%} & 7.8G & 64.0tokens/ms & 64.0M & 32.9M \\
    \bottomrule
    \end{tabular}
    \end{adjustbox}
    \label{tab:10}
\end{table*}

\section{Pruned Structure Distribution}
\label{app:6}

\begin{figure*}[ht]
    \centering
    \subfigure[12-layer BERT\textsubscript{\sf base}.]{
      \includegraphics[width=0.97\textwidth]{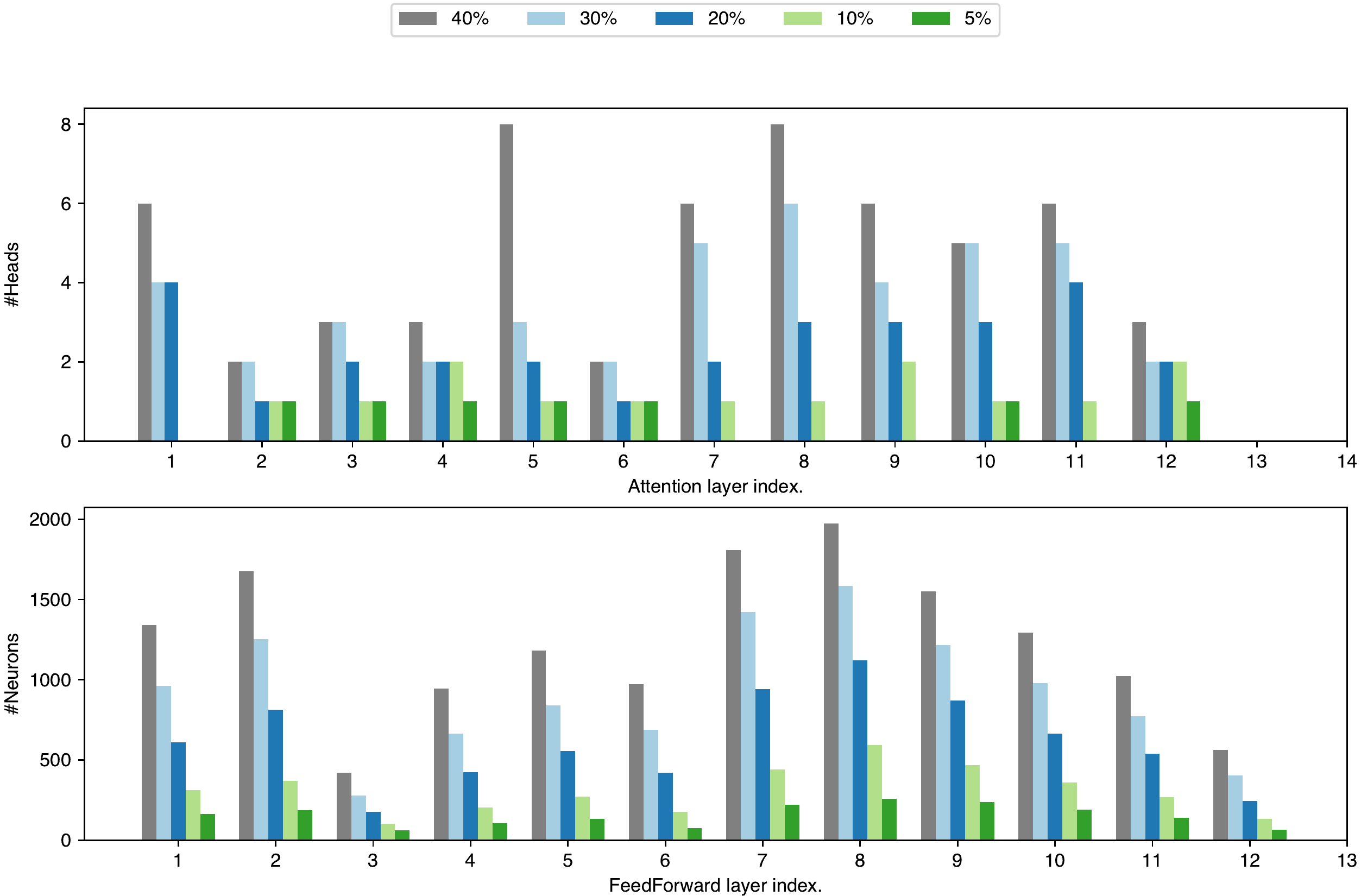}
      \label{fig:4a}
    }
    \subfigure[24-layer EncT5\textsubscript{\sf xl}. Layer indices lager than 24 denote modules from the one-layer decoder (i.e., two more attention modules and one more feed-forward modules).]{
      \includegraphics[width=0.99\textwidth]{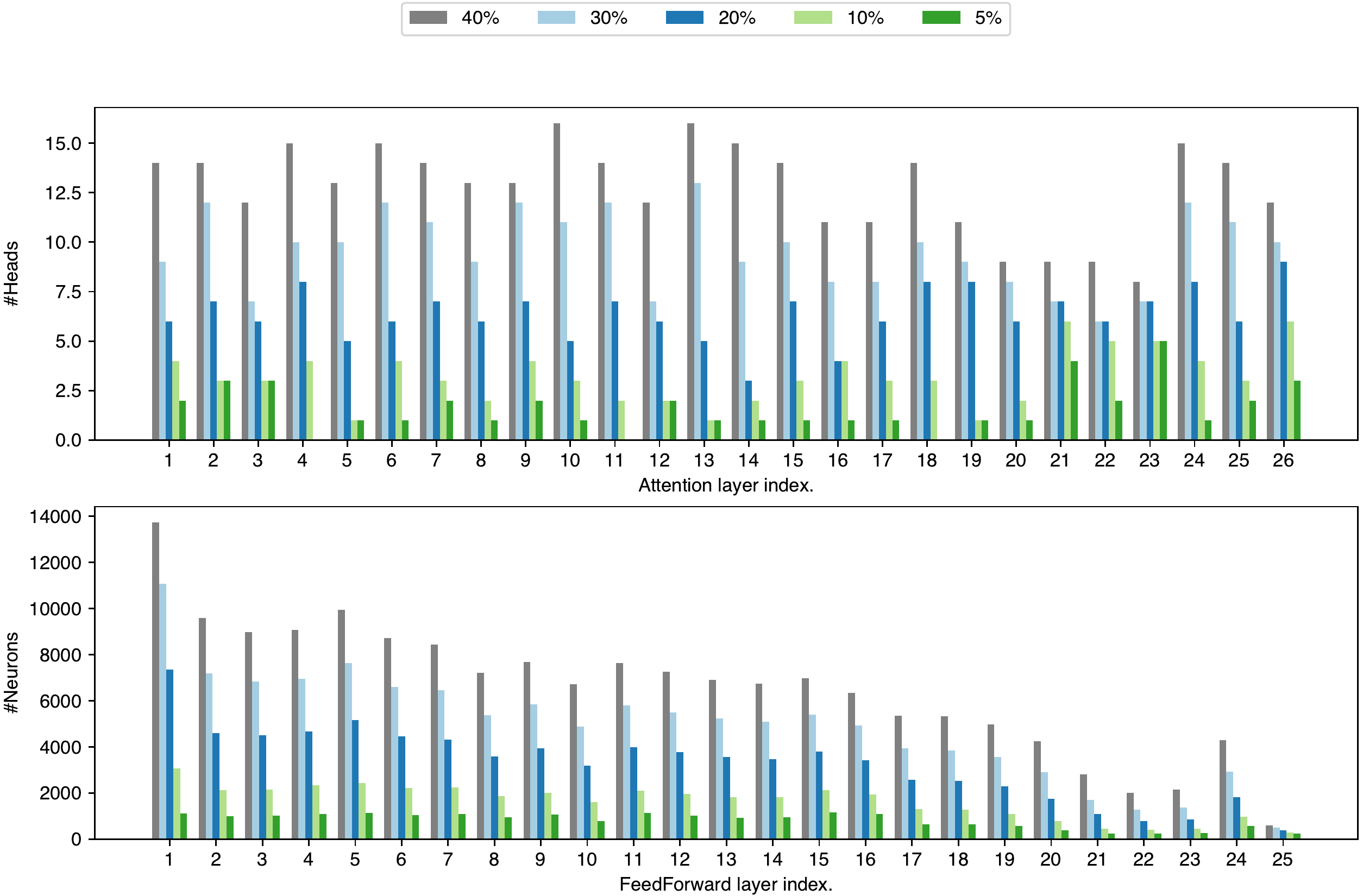}
      \label{fig:4b}
    }
    \caption{The distribution of example pruned structures. The structures are derived with MRPC dataset.}
    \label{fig:4}
\end{figure*}

We give the distribution of example pruned structures in Figure~\ref{fig:4}, which exactly show what pruned LMs consist of. While pruned BERT\textsubscript{\sf base} tends to preserve bottom and middle layers, pruned EncT5\textsubscript{\sf xl} tends to preserve bottom layers. Meanwhile, neurons in feed-forward layers are more likely to be pruned than heads in attention layers, owing to the centrality of the attention module within an transformer layer.

\section{Results upon \texorpdfstring{BERT\textsubscript{\sf large}}{BERT large}}
\label{app:7}

We show extended results of \textsc{MiniDisc} on BERT\textsubscript{\sf large} for readers' interest in Table~\ref{tab:11}. Consistent patterns have been observed as in BERT\textsubscript{\sf base}.

\begin{table*}[ht]
    \centering
    \caption{The results of task-specific distillation upon BERT\textsubscript{\sf large}.}
    \begin{adjustbox}{width=0.55\textwidth,center}
    \begin{tabular}{lrccccc}
    \toprule
      Method & FLOPs & SST-2 & MRPC & STS-B & RTE & Average \\
    \midrule
      BERT\textsubscript{\sf base} & 10.9G & 93.8 & 91.5 & 87.1 & 71.5 & 86.0 \\
    \midrule
      $\mathcal{L}_{\sf TSD}$\textsubscript{\sf 10\%} & 1.1G & 88.8 & 87.8 & 84.0 & 66.4 & 81.8 \\
      \textsc{MaxiDisc}\textsubscript{\sf 10\%} & 1.1G & 89.0 & 88.2 & 84.8 & 66.8 & 82.2 \\
      \textsc{MiniDisc}\textsubscript{\sf 10\%} & 1.1G & 89.1 & 88.4 & 85.4 & 68.2 & 82.7 \\
    \midrule
      $\mathcal{L}_{\sf TSD}$\textsubscript{\sf 5\%} & 0.5G & 85.4 & 85.5 & 83.9 & 63.2 & 79.5 \\
      \textsc{MaxiDisc}\textsubscript{\sf 5\%} & 0.5G & 86.1 & 87.0 & 84.1 & 65.7 & 80.7 \\
      \textsc{MiniDisc}\textsubscript{\sf 5\%} & 0.5G & 86.9 & 87.6 & 84.8 & 66.8 & 81.5 \\
    \midrule
      BERT\textsubscript{\sf large} & 38.7G & 94.2 & 92.5 & 90.1 & 75.5 & 88.1 \\
    \midrule
      $\mathcal{L}_{\sf TSD}$\textsubscript{\sf 10\%} & 3.9G & 90.4 & 88.1 & 87.0 & 66.1 & 82.9 \\
      \textsc{MaxiDisc}\textsubscript{\sf 10\%} & 3.9G & 90.6 & 88.9 & 87.1 & 67.2 & 83.4 \\
      \textsc{MiniDisc}\textsubscript{\sf 10\%} & 3.9G & 90.5 & 88.8 & 87.8 & 66.1 & 83.3 \\
    \midrule
      $\mathcal{L}_{\sf TSD}$\textsubscript{\sf 5\%} & 1.9G & 89.2 & 85.7 & 85.8 & 61.4 & 80.5 \\
      \textsc{MaxiDisc}\textsubscript{\sf 5\%} & 1.9G & 90.4 & 86.0 & 85.7 & 62.8 & 81.2 \\
      \textsc{MiniDisc}\textsubscript{\sf 5\%} & 1.9G & 89.6 & 87.4 & 87.3 & 61.4 & 81.4 \\
    \midrule
      EncT5\textsubscript{\sf xl} & 155.9G & 96.9 & 95.1 & 92.3 & 88.5 & 93.2 \\
    \midrule
      $\mathcal{L}_{\sf TSD}$\textsubscript{\sf 10\%} & 15.6G & 94.5 & 90.2 & 87.4 & 67.5 & 84.9 \\
      \textsc{MaxiDisc}\textsubscript{\sf 10\%} & 15.6G & 94.6 & 90.5 & 88.0 & 70.4 & 85.9 \\
      \textsc{MiniDisc}\textsubscript{\sf 10\%} & 15.6G & 94.6 & 91.5 & 87.8 & 72.2 & 86.5 \\
    \midrule
      $\mathcal{L}_{\sf TSD}$\textsubscript{\sf 5\%} & 7.8G & 92.9 & 88.0 & 83.4 & 58.8 & 80.8 \\
      \textsc{MaxiDisc}\textsubscript{\sf 5\%} & 7.8G & 93.0 & 88.0 & 83.9 & 67.5 & 83.1 \\
      \textsc{MiniDisc}\textsubscript{\sf 5\%} & 7.8G & 93.8 & 89.8 & 85.3 & 64.6 & 83.4 \\
    \bottomrule
    \end{tabular}
    \end{adjustbox}
    \label{tab:11}
\end{table*}

\section{Results of Small-scale Distillation}
\label{app:8}

When \textsc{MiniDisc} is applied to small MiniLM\textsubscript{\sf 12;384H} and BERT\textsubscript{\sf mini} as shown in Table~\ref{tab:12}, \textsc{MiniDisc} can reversely affect the performance of conventional distillation. Contrarily, \textsc{MaxiDisc} can still improve or at least retain the performance. However, it is less necessary to compress small LMs. 

\begin{table*}[ht]
    \centering
    \caption{The results of task-specific distillation upon small LMs.}
    \begin{adjustbox}{width=0.8\textwidth,center}
    \begin{tabular}{lrcccccccc}
    \toprule
      Method & FLOPs & SST-2 & MRPC & STS-B & QQP & MNLI-m/mm & QNLI & RTE & Average \\
    \midrule
      MiniLM\textsubscript{\sf 12L;384H} & 2.72G & 92.1 & 90.9 & 88.6 & 87.2 & 83.0/83.3 & 90.7 & 72.9 & 86.1 \\
    \midrule
      $\mathcal{L}_{\sf TSD}$\textsubscript{\sf 10\%} & 0.26G & 87.8 & 87.1 & 85.6 & 84.3 & 77.2/78.4 & 84.8 & 66.4 & 81.5 \\
      \textsc{MaxiDisc}\textsubscript{\sf 10\%} & 0.26G & 88.2 & 88.2 & 86.3 & 84.7 & 77.8/79.2 & 85.2 & 65.7 & 81.9 \\
      \textsc{MiniDisc}\textsubscript{\sf 10\%} & 0.26G & 87.6 & 86.0 & 86.5 & 84.4 & 77.8/78.6 & 84.4 & 64.6 & 81.3 \\
    \midrule
      BERT\textsubscript{\sf mini} & 0.60G & 87.5 & 86.4 & 85.3 & 85.0 & 76.1/77.2 & 84.5 & 66.8 & 81.1 \\
    \midrule
      $\mathcal{L}_{\sf TSD}$\textsubscript{\sf 10\%} & 0.04G &	83.3 & 83.8 & 81.6 & 81.6 & 66.3/71.4 & 82.7 & 58.8 & 76.2 \\
      \textsc{MaxiDisc}\textsubscript{\sf 10\%} & 0.04G & 83.8 & 84.1 & 80.7 & 82.0 & 66.4/71.6 & 82.9 & 58.1 & 76.2 \\
      \textsc{MiniDisc}\textsubscript{\sf 10\%} & 0.04G & 83.3 & 82.9 & 80.6 & 81.1 & 67.4/71.3 & 82.8 & 58.5 & 76.0 \\
    \bottomrule
    \end{tabular}
    \end{adjustbox}
    \label{tab:12}
\end{table*}

\section{Additional Task-specific Distillation for TinyBERT}
\label{app:9}

We compare TinyBERT with and without task-specific distillation as in Table~\ref{tab:13}. The results with task-specific distillation are retrieved from the original paper, since their augmented data is not publicly available. The results demonstrate that TinyBERT is largely supported with task-specific distillation and data augmentation for good performance.

\begin{table*}[ht]
    \caption{The results of TinyBERT with and without TSD.}
    \begin{adjustbox}{width=0.9\textwidth,center}
    \begin{tabular}{lrcccccccc}
    \toprule
      Method & FLOPs & SST-2 & MRPC & STS-B & QQP & MNLI-m/mm & QNLI & RTE & Average \\
    \midrule
      TinyBERT\textsubscript{\sf 4L;312H}~\citep{JiaoYSJCL0L20} & 0.6G & 88.5 & 87.9 & 86.6 & 85.6 & 78.9/79.2 & 87.3 & 67.2 & 82.7 \\
      \quad w/ TSD\&DA~\citep{JiaoYSJCL0L20} & 0.6G & 92.7 & 90.2 & 86.3 & 87.1 & 82.8/82.8 & 88.0 & 65.7 & 84.5 \\
      MiniLM\textsubscript{\sf 3L;384H}~\citep{WangBHDW21} & 0.7G & 89.1 & 89.1 & 86.6 & 85.4 & 77.8/78.4 & 87.2 & 66.1 & 82.5 \\
    \bottomrule
    \end{tabular}
    \end{adjustbox}
    \label{tab:13}
\end{table*}

\section{Negative Derivative-Tradeoff}
\label{app:10}

As mentioned in the main paper, although \textit{$\lambda$-tradeoff} is able to provide stable tradeoff measurement, it is dependent on the value of $\lambda$.
To eliminate this dependency, we design a new measure, negative derivative-tradeoff, which computes the negative derivative of performance to scale at each candidate scale as: $t_a = \lim_{\delta\to 0} \frac{-(m_{a+\delta} - m_a)}{s_{a+\delta} - s_a}$.
%where the minus sign is absorbed by $s_{a+\delta} - s_a$ since it is negative. 
In the discrete case, $t_{a_i} = \frac{- (m_{a_{i+1}} - m_{a_i})}{\Delta s_a} $. The idea of the measure is basically derived from saving the performance from a potentially significant drop. However, first-order estimation can lead to a high estimation variance and can be further tuned with second-order or so for better performance. The comparison results using \textit{$\lambda$-tradeoff} and \textit{ND-tradeoff} are shown in Table~\ref{tab:14}. It can be seen from the table that \textsc{MiniDisc}-ND also achieves comparable results.

\begin{table*}[ht]
    \centering
    \caption{The results of negative derivative-tradeoff upon BERT\textsubscript{\sf base}.}
    \begin{adjustbox}{width=0.60\textwidth,center}
    \begin{tabular}{lrccccc}
    \toprule
      Method & FLOPs & SST-2 & MRPC & STS-B & RTE & Average \\
    \midrule
      BERT\textsubscript{\sf base} & 10.9G & 93.8 & 91.5 & 87.1 & 71.5 & 86.0 \\
    \midrule
      $\mathcal{L}_{\sf TSD}$\textsubscript{\sf 10\%} & 1.1G & 88.8 & 87.8 & 84.0 & 66.4 & 81.8 \\
      \textsc{MaxiDisc}\textsubscript{\sf 10\%} & 1.1G & 89.0 & 88.2 & 84.8 & 66.8 & 82.2 \\
      \textsc{MiniDisc}-$\lambda$\textsubscript{\sf 10\%} & 1.1G & 89.1 & 88.4 & 85.4 & 68.2 & 82.7 \\
      \textsc{MiniDisc}-ND\textsubscript{\sf 10\%} & 1.1G & 89.8 & 87.9 & 85.4 & 66.4 & 82.4 \\
    \midrule
      $\mathcal{L}_{\sf TSD}$\textsubscript{\sf 5\%} & 0.5G & 85.4 & 85.5 & 83.9 & 63.2 & 79.5 \\
      \textsc{MaxiDisc}\textsubscript{\sf 5\%} & 0.5G & 86.1 & 87.0 & 84.1 & 65.7 & 80.7 \\
      \textsc{MiniDisc}-$\lambda$\textsubscript{\sf 5\%} & 0.5G & 86.9 & 87.6 & 84.8 & 66.8 & 81.5 \\
      \textsc{MiniDisc}-ND\textsubscript{\sf 5\%} & 0.5G & 86.8 & 86.0 & 84.9 & 66.8 & 81.1 \\
    \bottomrule
    \end{tabular}
    \end{adjustbox}
    \label{tab:14}
\end{table*}

\section{Varying Schedules for EncT5}
\label{app:11}

\begin{figure*}[ht]
    \centering
    \includegraphics[width=0.8\textwidth]{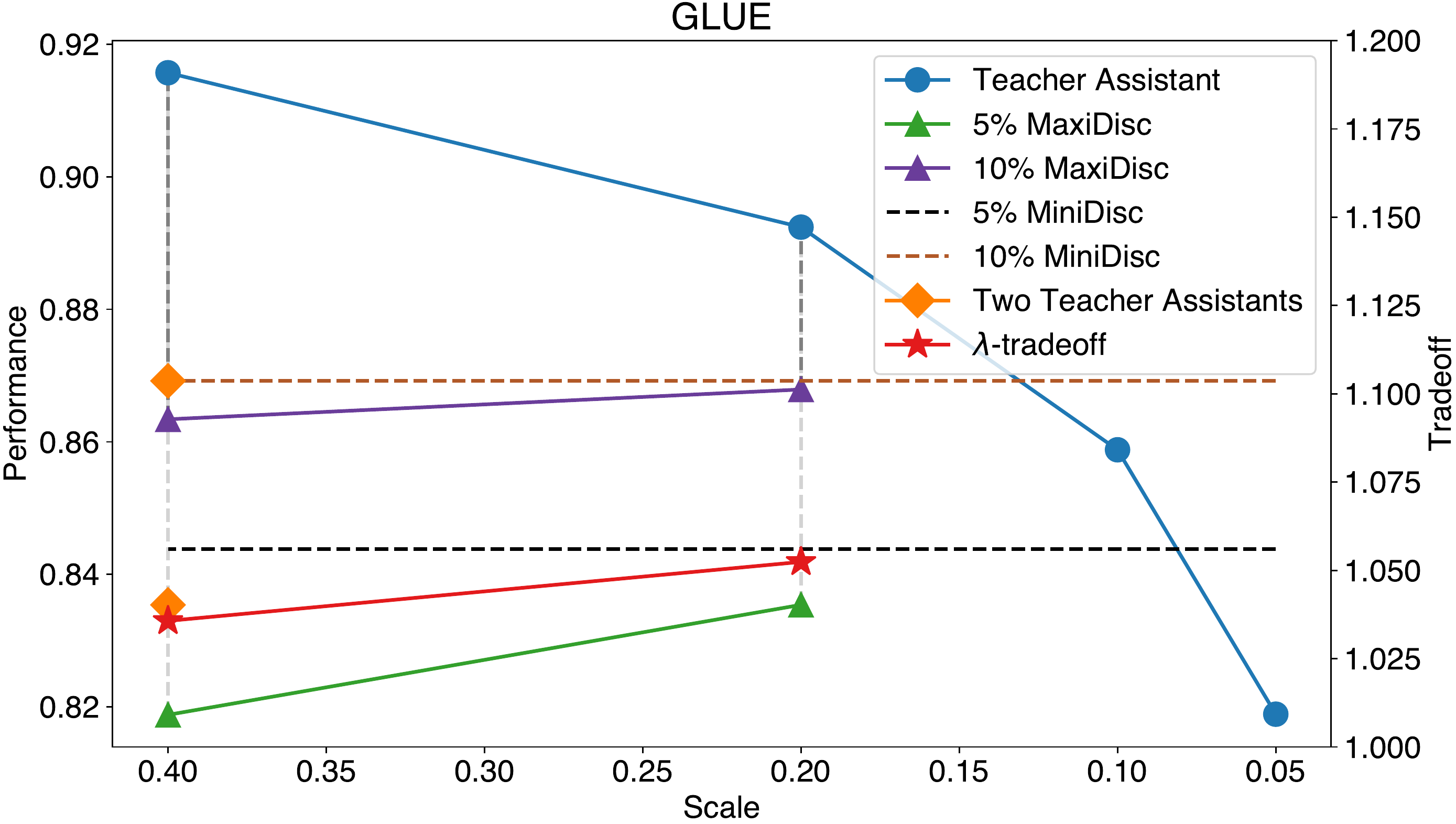}
    \caption{Performance comparisons among various schedules for EncT5. The dots represent performance variations using either one or two teacher assistants for \textsc{MaxiDisc}. The triangles represent performance resulting from \textsc{MiniDisc} using one teacher assistant. The rectangles represent performance resulting from \textsc{MiniDisc} using two teacher assistants.}
    \label{fig:5}
\end{figure*}

Performance variations among possible schedules for EncT5 are displayed in Figure~\ref{fig:5}, where the existence of scale-performance tradeoff and sufficiency of one teacher assistant can be verified.

\section{Residual Distillation}
\label{app:12}

The results in Table~\ref{tab:x} showcase that the follow-up action is at least a no-harm trick. 

\begin{table*}[t]
    \centering
      \caption{The results of residual distillation upon distilling BERT\textsubscript{\sf base} to BERT\textsubscript{\sf 10\%}.}
      \begin{adjustbox}{width=0.36\textwidth,center}
      \begin{tabular}{lcc}
      \toprule
        Method & MRPC & QQP \\
        \midrule
        $\mathcal{L}_{\sf TSD}$\textsubscript{\sf 10\%} & 87.8 & 84.6 \\
        \textsc{MiniDisc}\textsubscript{\sf 10\%} & 88.4 & 84.9 \\
        \quad w/ residual distillation & 88.4 & 85.1 \\
      \bottomrule
      \end{tabular}
      \end{adjustbox}
      \label{tab:x}
\end{table*}

\end{document}